\documentclass{article} 
\usepackage{iclr2027_conference,times}

\usepackage{amsmath,amsfonts,bm}

\def\eqref#1{equation~\ref{#1}}

\def\1{\bm{1}}

\DeclareMathAlphabet{\mathsfit}{\encodingdefault}{\sfdefault}{m}{sl}
\SetMathAlphabet{\mathsfit}{bold}{\encodingdefault}{\sfdefault}{bx}{n}

\usepackage{amsmath}
\usepackage{amssymb}
\usepackage{mathtools}
\usepackage{amsthm}
\usepackage{booktabs}
\usepackage{listings}
\usepackage[table]{xcolor}
\usepackage{xspace}
\usepackage{hyperref}
\usepackage{url}
\usepackage[capitalize,noabbrev]{cleveref}
\usepackage{algorithm}
\usepackage{algpseudocode}
\usepackage{graphicx}
\usepackage{bbm}
\usepackage{enumitem}
\usepackage{adjustbox} 
\usepackage[most]{tcolorbox}
\usepackage{wrapfig}
\usepackage{subcaption}
\usepackage{caption}
\theoremstyle{plain}

\theoremstyle{definition}

\theoremstyle{remark}

\title{DUET: Co-Evolving Solver and Grader Agents}

\author{
Fengyu Gao$^{1}$\thanks{Work done during an internship at Microsoft.}
\quad
Sourav Pal$^{2}$
\quad
Austin Z. Henley$^{2}$
\quad
Arjun Radhakrishna$^{2}$
\quad
Gustavo Soares$^{2}$
\\[0.8em]
$^{1}$University of Virginia, Charlottesville, United States
\\
$^{2}$Microsoft, Redmond, United States
\\[0.8em]
\texttt{wan6jj@virginia.edu}
\\
\texttt{\{palsourav, austin.henley, arradha, Gustavo.Soares\}@microsoft.com}
}

\iclrfinalcopy 
\begin{document}

\maketitle

\begin{abstract}

Agentic workflows are increasingly used across domains such as technology, finance, and enterprise operations.
As these agents become more widely deployed, continually improving them becomes increasingly important.
This raises an immediate challenge: \textit{How should the agent evolve?}
This evolution requires effective evaluation that can assess outcomes and provide useful feedback for optimization.
As the agent evolves, its behaviors and failure modes may also change, making a fixed evaluator increasingly inadequate.
Another fundamental question: \textit{How should we evaluate an evolving agent?}
These two challenges are inherently coupled; changes in agent behavior can expose limitations of the current evaluator, while a stronger evaluator provides more informative feedback for improving the agent.
Motivated by this interaction, we introduce \textbf{DUET}, a framework that jointly optimizes a solver agent and a grader agent to improve both.
DUET iteratively selects training tasks, executes them with the solver, evaluates the resulting outcomes with the grader, and uses a tool-using update module to revise the solver and the grader, alternating between the two across rounds.
By updating the grader within the optimization loop, DUET turns evaluation from a fixed source of feedback into a first-class optimization objective that adapts alongside the solver.
Experiments across four agent benchmarks show that DUET improves both solver and grader performance and consistently outperforms baselines that optimize the solver with a fixed grader.

\end{abstract}

\section{Introduction}
\label{sec:intro}

Large language model (LLM) agents are becoming increasingly capable of tool use, reasoning, interaction, and long-horizon decision making \citep{liu2024agentbench,ma2024agentboard}, enabling applications such as spreadsheet editing \citep{ma2024spreadsheetbench,zhu2026spreadsheetbench}, code debugging \citep{jimenez2024swe,yang2024swe}, and multi-step business workflows \citep{drouin2024workarena}.
A growing line of work seeks to improve these agents by optimizing components such as system prompts and skills \citep{wang2024promptagent,yuksekgonul2024textgrad,agrawal2025gepa,yang2026skillopt}.
These methods iteratively execute the agent, evaluate its outputs, and use the resulting scores or critiques to update its components.
We refer to the task-performing agent as the \emph{solver} and the agent evaluating its outcomes as the \emph{grader}.
The effectiveness of agent optimization therefore depends critically on the quality of the grader's feedback.

As the solver agent grows more capable, evaluating its outputs can become harder.
A weaker solver may fail in obvious ways, while a stronger solver may produce plausible outputs whose errors require closer inspection. 
The evaluation problem may therefore evolve as the solver improves.
This is especially challenging for {\it open-ended} agent tasks, where multiple execution paths or outcomes may be valid. 
For these tasks, quality can depend on the full execution process and resulting artifacts rather than a single reference answer.
In these settings, LLM-based grader agents are increasingly used to evaluate agent behavior and produce scores and natural-language critiques \citep{zheng2023judging,tan2025judgebench,mohammadi2025evaluation,hashemi2024llm}. 
As the solver evolves, the grader also needs to adapt its evaluation policy accordingly.

However, existing agent optimization methods typically keep the grader \textit{fixed} while using its scores or critiques to iteratively update the solver agent \citep{yuksekgonul2024textgrad,agrawal2025gepa,yang2026skillopt}. This design assumes that the grader remains an adequate source of feedback throughout optimization.
But as the solver evolves, it may exhibit new failure modes or finer quality differences that the initial grader does not capture \citep{tan2025judgebench,chen2026benchmarking}.
Its feedback may become less informative for guiding subsequent solver updates \citep{shen2026rethinking}.
This suggests that effective agent optimization may require the grader to evolve together with the solver.

Prior work on self-improving systems motivates treating evaluation quality as part of the optimization process.  Self-Rewarding Language Models use a model's own judgments to drive iterative preference training \citep{yuan2024self}, while Meta-Rewarding explicitly trains judgment quality using preferences over judgments \citep{wu2025meta}. 
We investigate this principle for tool-using agents, keeping underlying models fixed while jointly optimizing solver and grader prompts and skills. 
Intuitively, the two agents can provide learning signals for each other. As the solver evolves, it exposes new failure modes and finer distinctions that can help refine the grader. The refined grader provides more informative feedback for subsequent solver updates. 
This leads to a key question:

\begin{center}
\textit{Can we co-evolve a solver agent and a grader agent to improve both task execution and evaluation?}
\end{center}

To address this question, we introduce \textbf{DUET} (\textbf{D}ual \textbf{U}pdates for \textbf{E}valuation and \textbf{T}ask-solving), a framework that treats both the solver and the grader as optimization targets (\Cref{fig:duet}).
DUET proceeds iteratively by selecting a batch of training tasks, running the solver on them, evaluating the resulting outcomes with the grader, and using an update module to revise either agent, with solver and grader updates alternating across rounds.
This creates a reciprocal optimization loop in which solver executions provide evidence for improving the grader, while the improved grader provides better feedback for subsequent solver updates.

We evaluate DUET on four benchmarks spanning spreadsheet manipulation, multi-turn tool use, and retrieval-augmented question answering. 
Experiments show that DUET improves solver and grader performance across all four benchmarks.
Notably, starting from a minimal grader initialization, DUET achieves higher solver performance than SkillOpt \citep{yang2026skillopt} and GEPA \citep{agrawal2025gepa}, even when these baselines use the benchmark reference evaluator. 
These results suggest that DUET can reduce the need for carefully designed initial graders and reference answers.
Our contributions are summarized as follows:
\begin{figure*}[t]
\centering
\includegraphics[width=\textwidth]{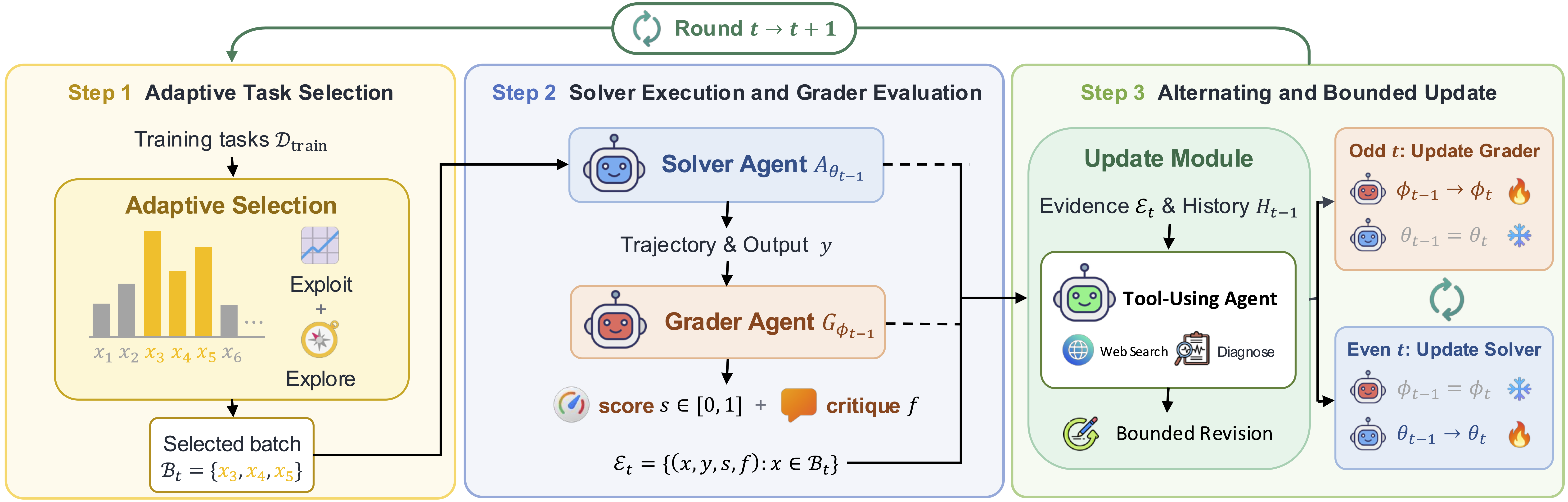}
\vspace{-15pt}
\caption{\small Overview of DUET. At each round, DUET first selects a training batch adaptively that balances exploitation and exploration. The solver then executes the selected tasks, and the grader evaluates the resulting trajectories and outputs to produce scores and critiques. Finally, the update module alternates between revising the grader and the solver, keeping the other fixed in each round. }
\vspace{-0.1in}
\label{fig:duet}
\end{figure*}

\begin{itemize}[topsep=0pt, left=0pt]
\item We introduce \textbf{DUET}, a framework for jointly optimizing a solver and a grader so that task execution and evaluation can improve together.

\item We develop a practical algorithm for solver--grader co-evolution with three key design elements: alternating solver--grader updates, adaptive task selection, and bounded component updates.

\item We demonstrate DUET's effectiveness across four agent benchmarks, different underlying models, and different optimizable agent components.

\end{itemize}

\section{Related Work}\label{sec:related_work}

\begin{figure*}[t]
    \centering
\includegraphics[width=\textwidth]{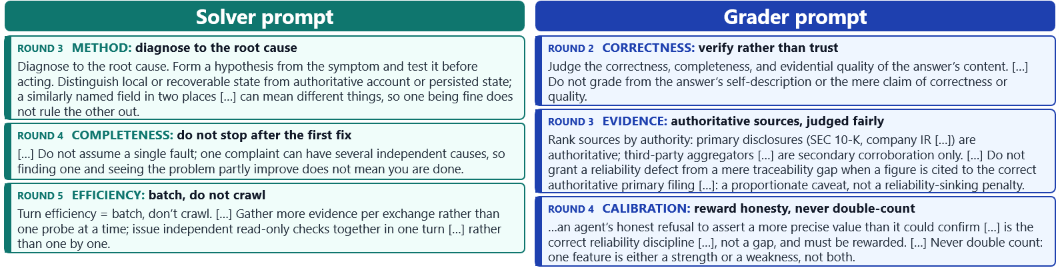}
    \vspace{-15pt}
    \caption{\small
    Representative solver and grader prompt updates produced by DUET during optimization.
    The solver develops reusable practices for root-cause diagnosis, completeness checking, and efficient tool use, while the grader develops practices for correctness verification, evidence assessment, and score calibration.
    }
    \label{fig:duete}
\end{figure*}

{\bf Agent Optimization.}
Recent work improves LLM agents by optimizing prompts, skills, and other agent components using task performance and evaluator feedback
\citep{zhou2023large,yuksekgonul2024textgrad,agrawal2025gepa,yang2026skillopt}.
For example, GEPA \citep{agrawal2025gepa} reflects on execution trajectories and evaluator feedback to diagnose failures and propose prompt updates, while maintaining a Pareto frontier of candidate prompts.
SkillOpt \citep{yang2026skillopt} uses scored rollouts to propose bounded edits to a persistent skill and accepts an update only when it improves performance on a held-out validation set.
We include GEPA and SkillOpt as representative baselines in our experiments.
Beyond single-agent optimization, AgentGrad \citep{chu2026agentgrad} investigates credit assignment in multi-agent prompt optimization, and MASPO \citep{wang2026maspo} jointly optimizes multiple solver agents in collaborative systems.
Across these lines of work, the evaluator that provides the optimization signal is typically fixed throughout optimization, including in GEPA and SkillOpt.
{\it DUET instead treats the grader itself as an optimization target, allowing task execution and evaluation to evolve together.}

{\bf LLM-Based Evaluation.}
LLMs are widely used to evaluate model outputs through LLM-as-a-judge approaches and rubric-based evaluators \citep{zheng2023judging,kim2024prometheus,kim2024prometheus2}.
Recent work develops agentic evaluators for assessing agents, where evaluation may require inspecting execution trajectories, intermediate actions, environment states, and generated artifacts \citep{zhuge2024agent,lin2026cuarewardbench}.
For example, Agent-as-a-Judge \citep{zhuge2024agent} uses an agentic evaluator to assess other agentic systems by examining the task-solving process and providing detailed feedback.
Such graders can provide richer evaluation signals than fixed automatic metrics, but designing them can require substantial domain expertise and manual effort to specify evaluation criteria, rubrics, reference materials, and judging instructions, especially when relevant failure modes are difficult to anticipate in advance \citep{pan2024human,shen2026rethinking}.
{\it DUET allows the grader to develop its evaluation policy from a minimal initialization, reducing reliance on manual grader design.}

{\bf Co-Evolution of Task Solving and Evaluation.} 
Recent work studies the co-evolution of task-solving and evaluation components \citep{li2026stale,zhang2026coevoskills}.
ECHO \citep{li2026stale} co-evolves a policy and a critic within an on-policy reinforcement learning framework, where an external outcome reward is used to train the policy and assess critic feedback.
Our method jointly optimizes the solver and the grader that directly provides the evaluation signal, without requiring a separate external reward model.
CoEvoSkills \citep{zhang2026coevoskills} co-evolves a skill generator and a surrogate verifier, where the generator iteratively refines agent skills and the verifier provides executable tests and diagnostic feedback.
Our grader agent can instead evaluate aspects of task quality that are not captured by executable tests alone.
We provide a broader discussion of related work and additional comparisons in Appendix~\ref{appx:related_work}.

\section{Problem Formulation}
\label{sec:problem}

We formalize agent optimization as the joint evolution of a solver and a grader.
Let \(\mathcal{D}\) denote a distribution over open-ended tasks \(x \in \mathcal{X}\).
A solver agent \(A_{\theta}\) is parameterized by an optimizable agent component \(\theta \in \Theta\), such as a system prompt, reusable skills, or a combination of these.
Given a task \(x\), the solver produces an execution outcome
\(y \sim P_{\theta}(\cdot \mid x)\), where \(y\) includes the agent's execution trajectory and the resulting output or artifacts.
A grader agent \(G_{\phi}\) is similarly parameterized by an optimizable agent component \(\phi \in \Phi\).
Given a task \(x\) and an execution outcome \(y\), the grader returns
\((s,f)=G_{\phi}(x,y)\), where \(s\) is a scalar score and \(f\) is a natural-language critique.
The underlying language models, tools, and execution environments are fixed. We optimize only the agent components \(\theta\) and \(\phi\).
This setting covers a broad range of solver and grader architectures, including prompt- or skill-based components \citep{agrawal2025gepa,yang2026skillopt}, rubric-based graders \citep{kim2024prometheus}, and tool-using agents \citep{zhuge2024agent}. Figure \ref{fig:duete} shows an example from DUET where the solver and grader were updated simultaneously.

We consider {\it reference-free training updates}, which do not use reference outputs or ground-truth quality labels.
In our default setting, a separate labeled validation set is used only for checkpoint selection and early stopping; we also evaluate a fixed-budget variant without validation-based selection or stopping (Appendix~\ref{appx:checkpoint_selection}).
This setting is well suited to \textbf{open-ended agent tasks}, where multiple outputs may be valid and task quality can depend on the full execution process and resulting artifacts, not just the final answer.
Constructing gold-standard references or labels for such tasks can be costly, and they may still fail to capture all aspects of a high-quality outcome, making reference-based optimization difficult to scale. Let \(q^{\star}(x,y)\) denote the true latent quality of outcome \(y\) for task \(x\).
The ideal solver would maximize expected true task quality:
\begin{equation}
\theta^{\star}
\in
\arg\max_{\theta\in\Theta}
\mathbb{E}_{x\sim\mathcal{D}}
\mathbb{E}_{y\sim P_{\theta}(\cdot\mid x)}
\left[q^{\star}(x,y)\right].
\end{equation}
For a given solver \(\theta\), an ideal grader should assign scores that faithfully reflect the true quality of its outcomes and provide informative critiques.
\begin{equation}
\phi^{\star}(\theta)
\in
\arg\min_{\phi\in\Phi}
\mathbb{E}_{x\sim\mathcal{D}}
\mathbb{E}_{y\sim P_{\theta}(\cdot\mid x)}
\left[
\mathcal{L}\!\left(G_{\phi}(x,y),q^{\star}(x,y)\right)
\right],
\end{equation}
where \(\mathcal{L}\) measures quality of the grader's evaluation, including score fidelity and the usefulness of its critique. Since \(q^{\star}\) is unavailable, neither objective can be optimized directly. Moreover, as the solver evolves, the distribution of its outcomes changes, so the ideal grader \(\phi^{\star}(\theta)\) also changes. 
Our goal is to co-evolve solver component \(\theta\) and grader component \(\phi\) without access to \(q^{\star}\), reference answers, or ground-truth quality labels during training. The solver produces higher-quality outcomes and the grader provides more faithful and informative evaluations of the solver's evolving behavior.

\section{DUET Algorithm}
\label{sec:method}

We introduce DUET, a framework that jointly optimizes a solver agent and a grader agent through iterative task execution and evaluation. 
In each round, DUET selects a batch of training tasks, the solver executes the selected tasks, and the grader evaluates the resulting trajectories and outputs with a score and evidence-grounded critique. 
An update module uses this feedback and the execution artifacts to revise the optimizable component of either the solver or the grader, with the two agents updated in alternating rounds.
Training updates do not use reference answers or ground-truth labels.
A separate labeled validation set is used only for checkpoint selection and stopping. DUET starts from solver and grader components, $\theta_0$ and $\phi_0$, with an empty training history $H_0$.

\subsection{Step 1: Adaptive Task Selection}

At the beginning of round $t$, DUET selects a batch $\mathcal{B}_t$ of training tasks.
As in curriculum learning \citep{bengio2009curriculum}, the value of a task as a training signal can change during optimization, motivating DUET to adapt which tasks are selected.
DUET tracks how the grader score for the same task changes across consecutive executions to identify tasks that remain responsive to solver or grader updates.
A large score change suggests that recent solver or grader updates have changed how the task is executed or evaluated.
These score changes suggest that the task is still exposing new solver behaviors or grader weaknesses, so revisiting it may provide useful feedback. 
By contrast, if a task receives nearly the same evaluation across repeated executions, it may already be saturated or currently too difficult to improve, making it less likely to provide a useful signal at that stage of optimization.
DUET therefore {\it prioritizes tasks with larger recent score changes while retaining exploration} so that stable tasks can still be revisited.

To obtain the score history needed for this criterion, DUET first uses round-robin selection until each training task has been executed twice.
Let $s_i(k)\in [0,1]$ denote the normalized grader score for task $i$ on its $k$-th execution. 
We measure the change in score between two consecutive executions of the same task as 
\begin{equation}
    r_i(k)=\left|s_i(k)-s_i(k-1)\right|.
\end{equation}
To emphasize recent changes, DUET considers these score differences over a sliding window of the previous $\tau$ rounds.
Let $\mu_{i,\tau}(t)$ denote the average recent score change for task $i$, and let $N_{i,\tau}(t)$ denote the number of corresponding observations.

DUET then combines recent score changes with exploration using the Sliding-Window Upper Confidence Bound (SW-UCB) algorithm \citep{garivier2011upper}.
For each task $i$, we compute
\begin{equation}
u_i(t)
=
\begin{cases}
\mu_{i,\tau}(t)+
\sqrt{\frac{c\log(\min\{t-1,\tau\})}{N_{i,\tau}(t)}},
& N_{i,\tau}(t)>0,\\
+\infty,
& N_{i,\tau}(t)=0,
\end{cases}
\label{eq:swucb}
\end{equation}
where $c$ is a hyperparameter that controls the strength of exploration.
The first term favors tasks with larger recent score changes, while the second encourages exploration of tasks with fewer recent observations.
DUET then selects the tasks with the largest $u_i(t)$ values to form the batch $\mathcal{B}_t$.
Additional details of the task-selection procedure are provided in Appendix~\ref{appx:task_selection}.

\subsection{Step 2: Solver Execution and Grader Evaluation}

For each selected task $x\in\mathcal{B}_t$, the solver uses its current component $\theta_{t-1}$ to produce an outcome
\begin{equation}
y \sim P_{\theta_{t-1}}(\cdot\mid x),
\end{equation}
where $y$ includes the execution trajectory and the resulting output or artifacts.
The grader then evaluates $y$ using its current component $\phi_{t-1}$,
\begin{equation}
(s,f)=G_{\phi_{t-1}}(x,y),
\end{equation}
where $s$ is the scalar score, normalized to $[0,1]$, and $f$ is detailed feedback that identifies strengths and weaknesses and supports them with evidence from the execution and artifacts.
We collect the solver outcomes and grader evaluations for the current batch as the round evidence 
\begin{equation}
\mathcal{E}_t
=
\left\{
(x,y,s,f):x\in\mathcal{B}_t
\right\}.
\label{eq:round_evidence}
\end{equation}
We denote the training history from previous rounds by
\begin{equation}
H_{t-1}
=
\left(
(\mathcal{E}_j,\theta_{j-1},\phi_{j-1},\theta_j,\phi_j)
\right)_{j=1}^{t-1},
\end{equation}
which records the round evidence and the solver and grader components at each update.

\subsection{Step 3: Alternating and Bounded Updates}
Given the current round evidence $\mathcal{E}_t$ and training history $H_{t-1}$, DUET next updates either the solver or the grader. The two agents depend on each other during optimization: grader feedback guides solver updates, while solver executions provide the evidence used to update the grader. 
Updating both simultaneously would change both the solver behavior and the evaluation policy, making it difficult to attribute the effect of each update.
DUET therefore {\it alternates between solver and grader updates}, so that only one agent is revised in each round while the other remains fixed.
Let $z_{t-1}\in\{A_{\theta_{t-1}},G_{\phi_{t-1}}\}$ denote the agent selected for update at round $t$.
DUET updates the grader on odd-numbered rounds and the solver on even-numbered rounds.

The selected agent is revised by an update module.
Let $\omega_{t-1}\in\{\theta_{t-1},\phi_{t-1}\}$ denote its current optimizable component.
The update module contains a tool-using update agent that proposes a candidate revision $\widetilde{\omega}_t$ given $\omega_{t-1}$, the round evidence $\mathcal{E}_t$, and the training history $H_{t-1}$.
Its system prompt specifies how the available evidence should be analyzed to produce this revision.

For solver updates, the prompt instructs the update agent to use grader feedback as the primary signal, verify it against the execution artifacts, and identify a reusable execution habit that should be added, reinforced, or revised.
For grader updates, the prompt instructs the update agent to examine the grader's report and investigation together with the execution artifacts to identify evaluation practices that should be added, reinforced, or revised, including rubric construction, artifact verification, score calibration, and evidence grounding.
Both prompts instruct the update agent to use $H_{t-1}$ to assess previous changes based on how performance changes when the same tasks are revisited.
The update agent can build on changes that helped and revise those that did not.
Before adding or modifying a habit or evaluation practice, the update agent is also instructed to search the web for relevant current practices.
The update agent then makes the smallest revision supported by the evidence, or leaves the component unchanged when no revision is justified.
The prompts also discourage task-specific rules to reduce overfitting.
The system prompts are provided in Appendix~\ref{appx:updater_prompts}.

To control the size of each revision, DUET uses an edit budget as a textual learning rate \citep{yang2026skillopt}.
The budget limits the number of discrete changes allowed in each update and is gradually reduced over optimization, allowing larger revisions early and smaller refinements later.
We follow the bounded-update procedure from \citet{yang2026skillopt} to obtain the updated component $\omega_t$ from $\omega_{t-1}$ and $\widetilde{\omega}_t$, with details provided in Appendix~\ref{appx:bounded_update}.
The resulting $\omega_t$ becomes the updated component of the selected agent, while the other agent remains unchanged.
$(\mathcal{E}_t,\theta_{t-1},\phi_{t-1},\theta_t,\phi_t)$ is appended to the training history to form $H_t$.

After each round, DUET evaluates the updated solver and grader on a separate labeled validation set and tracks their best checkpoints independently.
Optimization stops when neither validation metric improves for a fixed number of rounds.
We also obtain comparable performance with a fixed training budget and no validation-based checkpoint selection or early stopping (Appendix~\ref{appx:checkpoint_selection}). The complete DUET procedure is summarized in \Cref{alg:duet}.

\begin{figure}[h]
\centering

\noindent\rule{\linewidth}{0.4pt}

\vspace{-0.6em}
\captionof{algorithm}{DUET}
\label{alg:duet}
\vspace{-0.7em}

\noindent\rule{\linewidth}{0.4pt}

\vspace{0.2em}
\small
\begin{algorithmic}[1]

\State {\bf Input:} $\mathcal{D}_{\mathrm{train}}$, $\mathcal{D}_{\mathrm{val}}$,
initial solver $\theta_0$, initial grader $\phi_0$; initialize
$H_0\gets\emptyset$ and best checkpoints
\label{alg:duet:init}

\For{$t=1,2,\ldots$}
\label{alg:duet:round}

    \If{warm-up}
    \label{alg:duet:warmup}
        \State $\mathcal{B}_t \gets
        \textsc{RoundRobin}(\mathcal{D}_{\mathrm{train}})$
        \label{alg:duet:roundrobin}
    \Else
        \State $\mathcal{B}_t \gets$ tasks with largest $u_i(t)$
        from Eq.~\ref{eq:swucb}
        \label{alg:duet:adaptive}
    \EndIf

    \For{$x \in \mathcal{B}_t$}
    \label{alg:duet:rollout-loop}
        \State $y \sim P_{\theta_{t-1}}(\cdot \mid x)$
        \label{alg:duet:solver}
        \State $(s,f) \gets G_{\phi_{t-1}}(x,y)$
        \label{alg:duet:grader}
    \EndFor

    \State $\mathcal{E}_t \gets
    \{(x,y,s,f):x\in\mathcal{B}_t\}$
    \label{alg:duet:evidence}

    \State $(z_{t-1},\omega_{t-1})\gets
    (G_{\phi_{t-1}},\phi_{t-1})$ if $t$ is odd,
    else $(A_{\theta_{t-1}},\theta_{t-1})$
    \label{alg:duet:select-role}

    \State $\widetilde{\omega}_t
    \gets
    \textsc{UpdateAgent}
    (z_{t-1},\omega_{t-1},\mathcal{E}_t,H_{t-1})$
    \label{alg:duet:updater}

    \State $\omega_t
    \gets
    \textsc{BoundedAdopt}
    (\omega_{t-1},\widetilde{\omega}_t)$
    \label{alg:duet:bounded}

    \State $(\theta_t,\phi_t)\gets
    (\theta_{t-1},\omega_t)$ if $t$ is odd,
    else $(\omega_t,\phi_{t-1})$
    \label{alg:duet:update-state}

    \State $H_t \gets H_{t-1} \oplus
    (\mathcal{E}_t,\theta_{t-1},\phi_{t-1},
    \theta_t,\phi_t)$
    \label{alg:duet:history}

    \State Evaluate $(\theta_t,\phi_t)$ on
    $\mathcal{D}_{\mathrm{val}}$ and update best checkpoints
    \label{alg:duet:validation}

    \If{no validation improvement for $p$ rounds}
    \label{alg:duet:stop}
        \State \textbf{break}
        \label{alg:duet:break}
    \EndIf

\EndFor

\State \Return best solver and grader checkpoints
\label{alg:duet:return}

\end{algorithmic}
\vspace{-8pt}
\noindent\rule{\linewidth}{0.4pt}
\end{figure}

\section{Experiments}

{\bf Benchmarks.} We evaluate on four benchmarks spanning spreadsheet manipulation, multi-turn tool use, and retrieval-augmented question answering: SpreadsheetBench \citep{ma2024spreadsheetbench}, GDPval \citep{patwardhan2026gdpval}, \(\tau^2\)-Bench \citep{barres2025tau}, and Finance Agent Benchmark \citep{microsoft2026financebenchmark}. 
We use SpreadsheetBench Verified, a curated, expert-validated subset of 400 tasks from the original 912-task SpreadsheetBench benchmark \citep{shortcut2025spreadsheetbenchverified}. Each task requires editing an existing Excel workbook according to a natural-language instruction.
For GDPval, we use all 60 tasks whose reference deliverables include an Excel workbook.
\(\tau^2\)-Bench contains 150 multi-turn customer-service tasks across airline, retail, and telecom domains, using a simulated user and domain-specific tools over a stateful environment.
Finance Agent Benchmark contains 126 long-form financial question-answering tasks that require web retrieval and citation-supported responses. 
Additional details are provided in Appendix~\ref{appx:benchmark}.

{\bf Agent and Tool Setup.}
For SpreadsheetBench and GDPval, we use a standalone Excel agent that edits real workbooks through Office.js, while the grader can inspect workbook contents and rendered outputs using code execution and screenshot tools.
For \(\tau^2\)-Bench, we use the benchmark's native multi-turn orchestrator and domain-specific tool environment, with the grader restricted to read-only environment inspection tools that do not expose reference labels.
For Finance Agent Benchmark, we use a web-search-enabled agent, while the grader has access to code execution and shell tools.
Full agent configurations are provided in Appendix~\ref{appx:agent_setup}.

{\bf Baselines.} 
We compare DUET with the unoptimized initialization \((A_0,G_0)\), where both agents use empty system prompts, as well as two prompt optimization methods, SkillOpt \citep{yang2026skillopt} and GEPA \citep{agrawal2025gepa}.
SkillOpt and GEPA optimize only the solver prompt and keep the grader fixed throughout training.
For both baselines, we use Claude Opus 4.8 for the solver and grader, matching the default DUET setting.

We evaluate the methods under two grader settings.
In the \emph{reference-free setting}, the fixed grader for SkillOpt and GEPA is the initial grader \(G_0\), while DUET starts from the same \(G_0\) and jointly optimizes both the solver and grader.
This setting models the challenge of bootstrapping evaluation for a new task collection, where task instructions and execution environments are available but reference answers and task-specific grading rubrics or prompts have not been prepared for training.
It tests whether useful evaluation practices can be developed alongside solver improvement, reducing the need for manual grader design before optimization begins.
In the \emph{oracle-grader setting}, the benchmark's reference evaluator \(G^\star\) is used during solver optimization.
We denote the corresponding baselines as SkillOpt (\(G^\star\)) and GEPA (\(G^\star\)).
We also evaluate an oracle variant of DUET, denoted DUET (\(G^\star\)), in which \(G^\star\) is fixed and only the solver is optimized.
This setting measures solver optimization when a task-specific reference evaluation signal is available.

Since SkillOpt and GEPA do not optimize the grader, we evaluate grader improvement by comparing DUET's learned grader against its initialization \(G_0\).
Additional baseline details are in Appendix~\ref{appx:baseline}.

{\bf DUET Setup.}
In our main experiments, DUET optimizes the system prompts of both the solver and grader, with Claude Opus 4.8 used for both agents.
The optimizable prompts are initialized empty except for the solver on $\tau^2$-Bench, which uses a minimal protocol prompt required by the benchmark interface.
Model weights, tools, and execution environments remain fixed during optimization.
We separately study alternative models, prompt initializations, and joint prompt-and-skill optimization in \Cref{tab:model_scale,tab:init,tab:skills}.

We implement the update agent described in Section~\ref{sec:method} using GitHub Copilot CLI with Claude Opus 4.8. We use an exploration coefficient of $c=0.01$ and the bounded-edit schedule described in Appendix~\ref{appx:bounded_update}, with sensitivity analyses in Appendix~\ref{appx:ablation_hyperparameter}.
Benchmark-specific batch sizes, window sizes, and stopping criteria are provided in Appendix~\ref{appx:parameters}.

{\bf Evaluation.}
We evaluate both the solver and the grader on held-out test data.
For the solver, we use each benchmark's reference evaluator.
On SpreadsheetBench, we report task accuracy, where a task is correct only if all evaluated cells match the reference workbook.
On GDPval, we report pairwise win rate against the human gold deliverable using the benchmark's pairwise reference grader.
On \(\tau^2\)-Bench, we report task success rate based on the benchmark-specified checks.
On Finance Agent Benchmark, we report the mean LLM-judge score across seven rubric dimensions.

For grader evaluation, we construct dedicated pairwise evaluation sets for all four benchmarks.
For GDPval and Finance Agent Benchmark, we generate multiple candidate outputs with Claude Opus 4.8 for each task and use the benchmark reference evaluator to establish their relative quality.
We retain pairs with reliable reference preferences and further select challenging cases where superficial cues favor the worse output.
For \(\tau^2\)-Bench and SpreadsheetBench, whose native evaluators provide only binary task outcomes, we generate two outputs per task using different underlying models
and retain pairs in which one output succeeds and the other fails.

Following prior work on LLM-based judge evaluation \citep{tan2025judgebench,lambert2025rewardbench}, we measure grader performance using pairwise agreement.
The grader scores each output independently, and receives credit 1 for preferring the reference-preferred output, 0.5 for a tie, and 0 for an inversion.
Additional evaluation details are available in Appendix~\ref{appx:eval}.

\subsection{Understanding the Performance of DUET}\label{sec:main_results}

In this section, we analyze the performance of DUET from both the solver and grader perspectives.
Our main results are presented in \Cref{tab:main}.

\begin{table*}[t]
\caption{\small
Solver and grader performance on the test sets using Claude Opus 4.8, with system prompts as the optimization target.
Solver metrics are rubric score for Finance Agent Benchmark, success rate for \(\tau^2\)-Bench, win rate for GDPval, and task accuracy for SpreadsheetBench.
Grader performance is measured by pairwise agreement.
\(A_0\) and \(G_0\) denote empty-prompt initialization, and \(G^\star\) denotes the benchmark reference evaluator.
Results show mean and standard deviation over 3 random seeds.
}
\label{tab:main}
\vspace{-10pt}
\begin{center}
\renewcommand{\arraystretch}{1.05}
\adjustbox{width=\textwidth}{
\footnotesize
\begin{tabular}{lcccc}
\toprule

\multicolumn{5}{c}{\textbf{Solver Performance}} \\

Method
& Finance Agent Benchmark
& \(\tau^2\)-Bench
& GDPval
& SpreadsheetBench \\
\midrule

\multicolumn{5}{l}{\textit{Reference-free setting}} \\

\(A_0\)
& $92.8\pm0.6$
& $80.3\pm2.1$
& $47.6\pm3.3$
& $72.6\pm0.7$ \\

SkillOpt
& $94.0\pm1.4$
& $84.7\pm6.4$
& $48.8\pm4.0$
& $81.6\pm1.1$ \\

GEPA
& $94.4\pm1.1$
& $83.3\pm5.8$
& $50.7\pm1.6$
& $82.8\pm1.4$ \\

\rowcolor{teal!10}
DUET
& $\bf{96.5\pm0.4}$
& $\bf{89.4\pm2.5}$
& $\bf{52.2\pm1.8}$
& $\bf{84.7\pm0.4}$ \\

\midrule

\multicolumn{5}{l}{\textit{Oracle-grader setting}} \\

SkillOpt (\(G^\star\))
& $95.7\pm0.4$
& $83.9\pm2.7$
& $47.1\pm2.4$
& $83.4\pm0.4$ \\

GEPA (\(G^\star\))
& $95.9\pm0.4$
& $82.0\pm4.1$
& $50.7\pm2.6$
& $83.0\pm2.0$ \\

\rowcolor{teal!10}
DUET (\(G^\star\))
& $\bf{96.0\pm0.2}$
& $\bf{86.1\pm2.7}$
& $\bf{54.9\pm2.4}$
& $\bf{84.7\pm1.0}$ \\

\midrule

\multicolumn{5}{c}{\textbf{Grader Performance}} \\

Method
& Finance Agent Benchmark
& \(\tau^2\)-Bench
& GDPval
& SpreadsheetBench \\
\midrule

\(G_0\)
& $67.9\pm1.1$
& $88.7\pm3.1$
& $68.4\pm1.0$
& $71.9\pm1.3$ \\

\rowcolor{teal!10}
DUET
& $\bf{75.0\pm2.9}$
& $\bf{89.3\pm2.3}$
& $\bf{78.5\pm6.1}$
& $\bf{72.8\pm1.9}$ \\

\bottomrule
\end{tabular}
}
\end{center}
\vspace{-10pt}
\end{table*}

\textit{RQ1: Can DUET improve solver performance compared with optimizing with a fixed grader?}
{\bf DUET consistently achieves stronger solver performance than fixed-grader prompt optimization across all four benchmarks.}
For example, on Finance Agent Benchmark, DUET achieves 96.5\% rubric score, improving over the unoptimized solver \(A_0\) at 92.8\%, SkillOpt at 94.0\%, and GEPA at 94.4\%.
More notably, it also exceeds SkillOpt and GEPA when they optimize against the benchmark reference evaluator $G^\star$, which reach 95.7\% and 95.9\%, respectively.
Similarly, on \(\tau^2\)-Bench, DUET reaches a success rate of 89.4\%, compared with 80.3\% for \(A_0\), 84.7\% for SkillOpt, and 83.3\% for GEPA, and 83.9\% and 82.0\% for their oracle-grader variants.

Our results highlight an advantage of DUET in reference-free open-ended tasks.
DUET does not require a strong grader at initialization, which is useful when reference answers or complete task-specific rubrics are unavailable.
Instead, the grader improves alongside the solver and can adapt its evaluation as the solver's behavior changes.
The fact that DUET can outperform fixed-grader methods even when they use the benchmark reference evaluator highlights the benefit of co-evolution.

\textit{RQ2: Can DUET improve the grader's evaluation capability?}
{\bf DUET improves grader evaluation capability while simultaneously optimizing the solver.}
On Finance Agent Benchmark, grader pairwise agreement improves from 67.9\% for the initial grader \(G_0\) to 75.0\% after DUET training.
On GDPval, agreement improves from 68.4\% to 78.5\%, a 10.1 percentage point gain.

For \(\tau^2\)-Bench and SpreadsheetBench, the reference evaluators provide only binary success or failure labels, so preference pairs can be constructed only when one output succeeds and the other fails.
Grader agreement therefore changes only slightly, from 88.7\% to 89.3\% on \(\tau^2\)-Bench and from 71.9\% to 72.8\% on SpreadsheetBench.
One possible explanation is that these pairs mainly capture coarse pass--fail differences that the initial grader already recognizes, while the subtler comparisons that could reveal improved grader judgment are absent from the benchmark.

We also report the inference cost of DUET compared with the baselines in Appendix~\ref{appx:cost}. {\bf DUET achieves stronger task performance with comparable token usage and numbers of LLM calls.} We also provide examples of prompt updates during optimization in Appendix~\ref{appx:prompt_evolution} and learned prompts and skills in Appendix~\ref{appx:DUET_prompts}.

\subsection{Understanding the Properties of DUET}\label{sec:property_DUET}

\begin{wraptable}{r}{0.38\textwidth}
    \vspace{-50pt}
    \caption{\small Success rate on \(\tau^2\)-Bench across optimization targets.}
    \vspace{-0.1in}
    \label{tab:skills}
    \begin{center}
    \renewcommand{\arraystretch}{1.2}
    \adjustbox{max width=\linewidth}{
    \begin{tabular}{lcc}
    \toprule
    Method & Prompt Only & Prompt + Skills \\
    \midrule
    SkillOpt & $84.7\pm{6.4}$ & $83.3\pm{2.2}$ \\
    GEPA & $83.3\pm{5.8}$ & $84.4\pm{7.1}$ \\
    \rowcolor{teal!10}
    DUET & $\bf{89.4\pm{2.5}}$ & $\bf{91.9\pm{4.1}}$ \\
    \bottomrule
    \end{tabular}
    }
    \end{center}
    \vspace{-0.1in}
\end{wraptable}

\textit{RQ3: Can DUET optimize agent components beyond the system prompt?} 
{\bf DUET extends beyond prompt optimization and can jointly optimize the system prompt and an agent skills library.}
On \(\tau^2\)-Bench, joint prompt-and-skill optimization achieves a 91.9\% success rate, improving over DUET with prompt-only optimization (89.4\%) and outperforming SkillOpt (83.3\%) and GEPA (84.4\%) (\Cref{tab:skills}).
One reason for this improvement is that \(\tau^2\)-Bench contains domain-specific subtasks, allowing DUET to encode specialized procedures as skills rather than incorporating all behavior into a single prompt.
Additional details are provided in Appendix~\ref{appx:skills}.

\begin{wraptable}{r}{0.38\textwidth}
\vspace{-0.2in}
\caption{\small Performance with different models on Finance Agent Benchmark.}
\vspace{-0.1in}
\label{tab:model_scale}
\begin{center}
\renewcommand{\arraystretch}{1.2}
\adjustbox{max width=\linewidth}{
\begin{tabular}{lcc}
\toprule

Method
& Claude Opus 4.8
& Claude Sonnet 4.6 \\

\multicolumn{3}{l}{\textbf{Solver Performance}} \\

$A_0$
& $92.8$
& $89.6$ \\

SkillOpt
& $94.0$
& $90.3$ \\

GEPA
& $94.4$
& $90.9$ \\

\rowcolor{teal!10}
DUET
& $\mathbf{96.5}$
& $\mathbf{91.3}$ \\

\midrule

\multicolumn{3}{l}{\textbf{Grader Performance}} \\

$G_0$
& $67.9$
& $58.3$ \\

\rowcolor{teal!10}
DUET
& $\mathbf{75.0}$
& $\mathbf{69.2}$ \\

\bottomrule
\end{tabular}
}
\end{center}
\vspace{-0.1in}
\end{wraptable}

\textit{RQ4: Does DUET remain effective with different underlying models?}
We additionally evaluate DUET with Claude Sonnet 4.6 on Finance Agent Benchmark, alongside the default Claude Opus 4.8 setting (\Cref{tab:model_scale}).
DUET achieves the highest solver performance while improving grader pairwise agreement.
We also evaluate DUET with GPT-5.6 Sol on $\tau_2$-Bench and SpreadsheetBench, where it improves $\tau_2$-Bench solver performance and SpreadsheetBench grader agreement. Full results are reported in Appendix~\ref{appx:gpt56}.

\begin{table*}[t]
    \caption{\small Effect of prompt initialization on Finance Agent Benchmark.}
    \vspace{-0.1in}
    \label{tab:init}
    \footnotesize
    \begin{center}
    \renewcommand{\arraystretch}{1.2}
    \adjustbox{max width=\linewidth}{
    \begin{tabular}{lcccc}
    \toprule
    Initialization 
    & Initial Solver 
    & \cellcolor{teal!10}DUET Solver 
    & Initial Grader 
    & \cellcolor{teal!10}DUET Grader \\
    \midrule
    Empty 
    & $92.8\pm{0.6}$ 
    & \cellcolor{teal!10}$\bf{96.5\pm{0.4}}$ 
    & $67.9\pm{1.1}$ 
    & \cellcolor{teal!10}$\bf{75.0\pm{2.9}}$ \\

    Informed 
    & $93.3\pm{0.2}$ 
    & \cellcolor{teal!10}$\bf{95.0\pm{0.5}}$ 
    & $68.2\pm{4.4}$ 
    & \cellcolor{teal!10}$\bf{73.3\pm{6.4}}$ \\

    Adversarial 
    & $91.1\pm{0.7}$ 
    & \cellcolor{teal!10}$\bf{95.8\pm{0.2}}$ 
    & $68.0\pm{2.0}$ 
    & \cellcolor{teal!10}$\bf{68.8\pm{1.4}}$ \\

    \bottomrule
    \end{tabular}
}
\end{center}
\end{table*}

\textit{RQ5: How does prompt initialization affect DUET?}
We evaluate DUET with empty, informed, and adversarial prompt initializations on Finance Agent Benchmark (\Cref{tab:init}).
DUET improves both solver and grader performance under all three initializations, including improving the adversarial solver from 91.1\% to 95.8\%.
Interestingly, DUET removes the harmful adversarial instructions in the first update, while retaining about 60\% of the informed solver prompt in the final prompt.
These results suggest that DUET can correct poor initializations while preserving useful prior structure.
Additional details are provided in Appendix~\ref{appx:init}.

\subsection{Validating the Design of DUET}\label{sec:validate_DEUT}

\begin{figure}[t]
    \centering

    \begin{subfigure}[t]{0.32\linewidth}
        \centering
        \includegraphics[width=\linewidth]{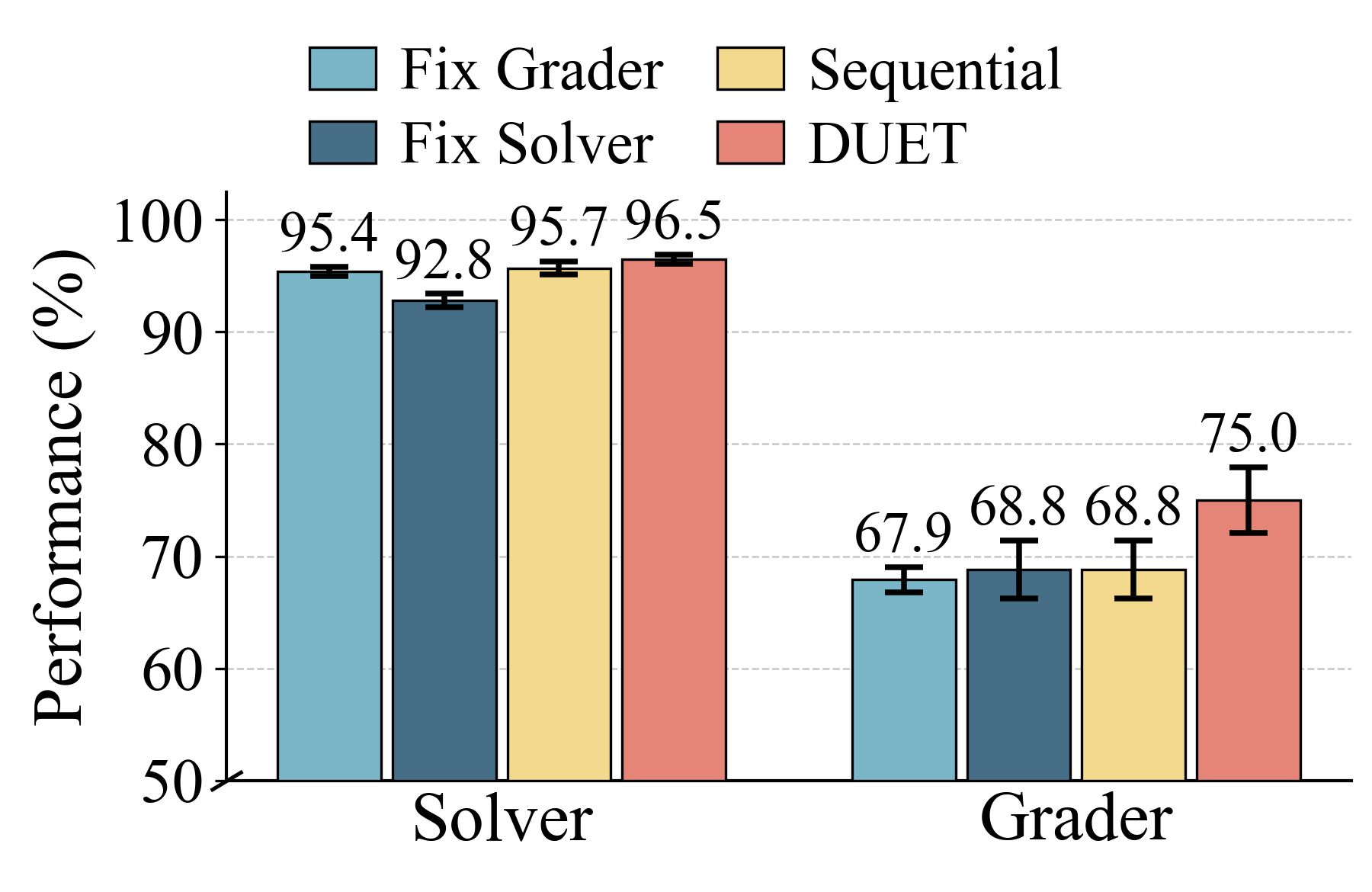}
        \vspace{-0.2in}
        \caption{Solver--grader co-evolution.}
        \label{fig:both_agents}
    \end{subfigure}
    \hfill
    \begin{subfigure}[t]{0.32\linewidth}
        \centering
        \includegraphics[width=\linewidth]{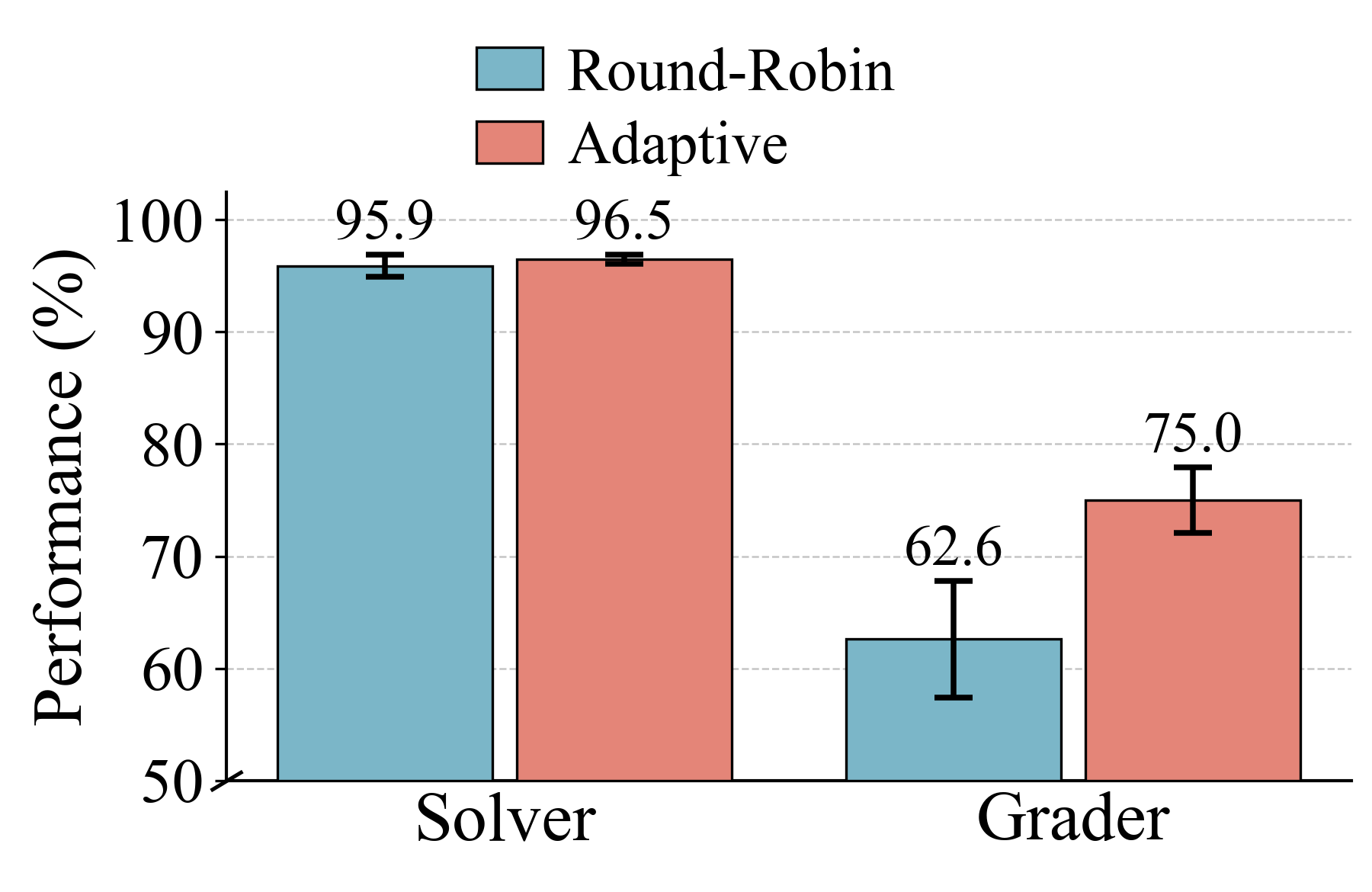}
        \vspace{-0.2in}
        \caption{Adaptive task selection.}
        \label{fig:curriculum}
    \end{subfigure}
    \hfill
    \begin{subfigure}[t]{0.32\linewidth}
        \centering
        \includegraphics[width=\linewidth]{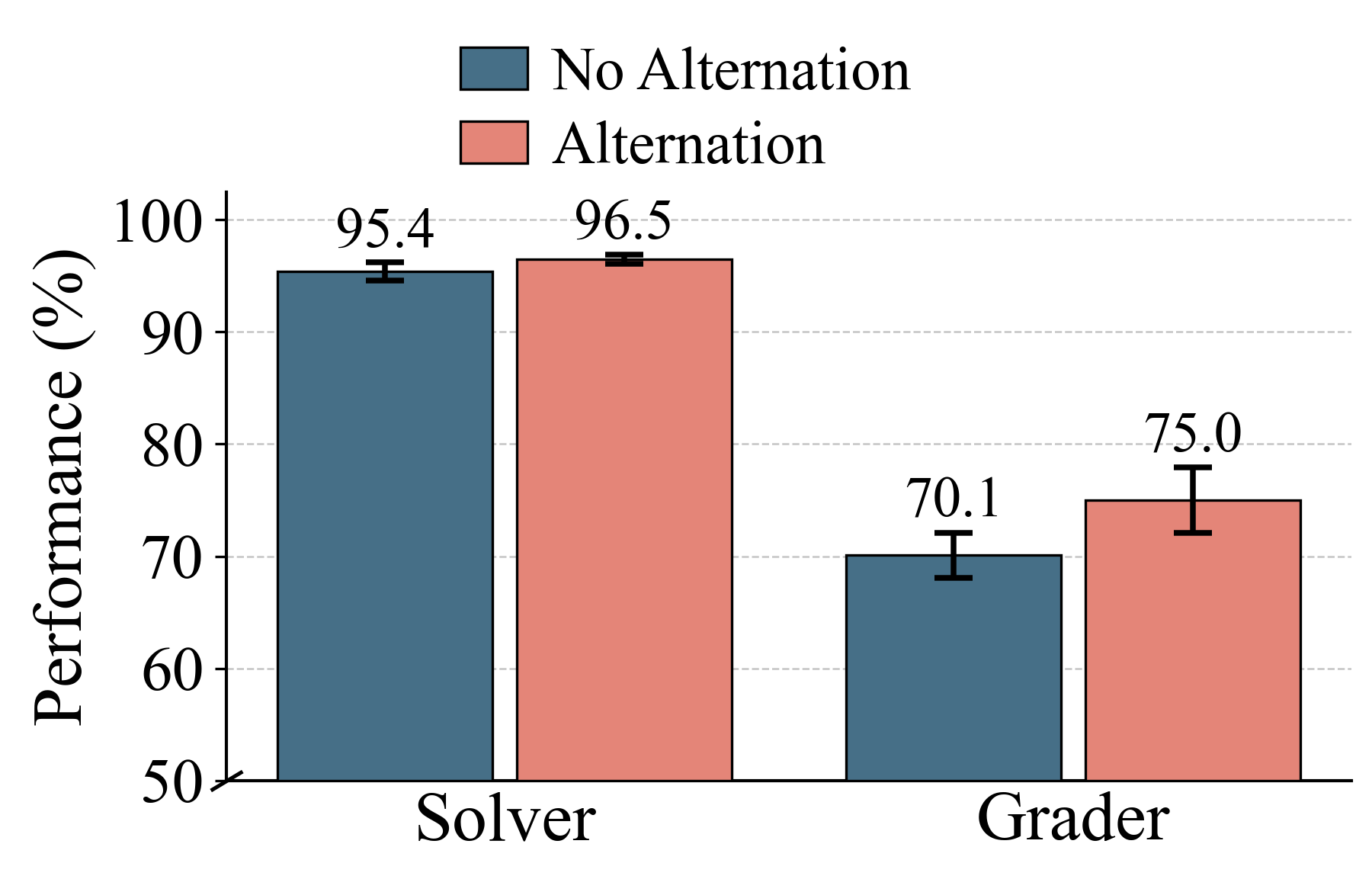}
        \vspace{-0.2in}
        \caption{Alternating updates.}
        \label{fig:alt}
    \end{subfigure}
\vspace{-0.1in}
    \caption{\small DUET ablations on Finance Agent Benchmark.}
    \label{fig:duet_ablations}
\end{figure}

\textit{RQ6: Why is co-evolving the solver and grader helpful?}
We compare DUET with three ablations on Finance Agent Benchmark: optimizing only the solver with a fixed grader, optimizing only the grader with a fixed solver, and a sequential variant that first optimizes the grader and then freezes it while optimizing the solver (\Cref{fig:both_agents}).
DUET achieves the strongest performance, reaching 96.5\% solver score and 75.0\% grader pairwise agreement.
When the solver is fixed, optimizing the grader improves pairwise agreement only from 67.9\% to 68.8\%.
This suggests that grader improvement benefits from an evolving solver that continues to expose new evaluation challenges.
Similarly, an evolving grader can identify new weaknesses and provide feedback for further solver improvement.

\textit{RQ7: Does adaptive task selection improve DUET?}
We compare DUET with and without adaptive task selection on Finance Agent Benchmark.
Without adaptive selection, tasks are selected in round-robin order.
As shown in \Cref{fig:curriculum}, adaptive task selection improves solver performance from 95.9\% to 96.5\% and grader pairwise agreement from 62.7\% to 75.0\%.
This suggests that revisiting tasks with changing evaluations provides more useful training signals.

\textit{RQ8: Do alternating solver and grader updates improve DUET?}
We compare DUET's alternating update scheme with a variant that updates both agents in every round on Finance Agent Benchmark.
As shown in \Cref{fig:alt}, alternating updates improve solver performance from 95.4\% to 96.5\% and grader pairwise agreement from 70.1\% to 75.0\%.
One advantage of alternation is clearer attribution: when only one agent is updated, changes in subsequent performance can be associated with that update while the other agent remains fixed.
When both are updated simultaneously, their effects are harder to disentangle.

\section{Conclusion}

In this work, we introduced DUET, a framework for jointly optimizing a solver agent and a grader agent so that task solving and evaluation can evolve together.
Across four benchmarks, DUET improves both solver and grader performance and outperforms baselines that optimize the solver with a fixed grader. 
These results suggest that, as agents evolve, evaluation should evolve alongside them rather than remain a fixed source of feedback.
One limitation of this work is that our agent evaluation relies on benchmark reference evaluators without human judgments, due to the cost of large-scale human evaluation.



\bibliography{iclr2027_conference}
\bibliographystyle{iclr2027_conference}

\appendix

\newpage

\section{Additional Related Work}\label{appx:related_work}

{\bf Agent Optimization.} A growing body of work improves LLM agents by optimizing components such as prompts and skills.
Prompt optimization methods typically use task performance and feedback to iteratively improve agent instructions \citep{zhou2023large,yang2024large,agarwal2024promptwizard,fernando2024promptbreeder,guo2024connecting,yuksekgonul2024textgrad,cheng2024trace,pryzant2023automatic,opsahl2024optimizing,agrawal2025gepa, sharma2026mpo}.
For example, GEPA \citep{agrawal2025gepa} reflects on execution trajectories and evaluator feedback to diagnose failures and propose prompt updates, while maintaining a Pareto frontier of candidate prompts.
We include GEPA as a representative baseline in our experiments.
AgentGrad \citep{chu2026agentgrad} investigates credit assignment in multi-agent prompt optimization by sequentially intervening on individual agents to identify which agent's prompt should be revised, then clustering and abstracting related textual gradients.
DUET instead studies reciprocal optimization of a solver and the grader that evaluates it, rather than credit assignment.
MASPO \citep{wang2026maspo} jointly optimizes task-performing agents in a collaborative multi-agent system using downstream-aware evaluation and evolutionary search, while DUET makes evaluation itself an optimization target.

Recent work also explores constructing and evolving skills from agent experience and task-relevant resources \citep{wang2025agent,wang2025inducing,fang2026memp,xia2026skillrl,ni2026trace2skill,alzubi2026evoskill,liu2026skillforge,shen2026skillfoundry,yang2026autoskill,wang2026skillx,qiu2026autorefine,mi2026procmem,wu2025evolver,yang2026skillopt}.
For example, SkillOpt \citep{yang2026skillopt} uses scored rollouts to propose bounded edits to a persistent skill and accepts an update only when it improves performance on a held-out validation set.
We also include SkillOpt as a baseline in our experiments.
Across these lines of work, the evaluator that provides the optimization signal is typically fixed throughout optimization, including in GEPA and SkillOpt.
{\it Our work instead jointly optimizes the solver and grader, allowing both task execution and evaluation to improve over time.}

{\bf LLM-Based Evaluation.}
LLMs have been widely used to evaluate the quality of model-generated outputs, providing a scalable way to assess complex aspects of output quality that are difficult to capture with fixed automatic metrics \citep{liu2023g,zheng2023judging,kim2024prometheus,kim2024prometheus2,mcaleese2024llm,zhu2025judgelm,chan2024chateval,zhao2025auto,lu2023llmscore,zeng2024evaluating}.
Building on this, recent work has explored AI judging systems that go beyond a single LLM call by coordinating multiple models, using tools, or performing multi-step evaluation \citep{chan2024chateval,chen2026multi,hu2026multi,shi2026aj}.
This is particularly useful for evaluating agents, where judging task performance may require considering not only the final output but also execution trajectories, intermediate actions, environment states, and generated artifacts \citep{zhuge2024agent,pan2024autonomous,ma2024agentboard,chen2024t,men2025agent,lin2026cuarewardbench}.
For example, Agent-as-a-Judge \citep{zhuge2024agent} uses an agentic evaluator to assess other agentic systems by examining the task-solving process and providing detailed feedback.
However, designing a reliable grader agent can require substantial domain expertise and manual effort to specify evaluation criteria, rubrics, reference materials, and judging instructions, especially when relevant failure modes are difficult to anticipate in advance \citep{pan2024human,shen2026rethinking}.
{\it DUET allows the grader to develop its evaluation policy from a minimal initialization, reducing reliance on extensive manual grader design.}

{\bf Co-Evolution of Task Solving and Evaluation.} There is a line of work on self-improvement that uses self-generated feedback to improve subsequent outputs or behavior \citep{madaan2023self,shinn2023reflexion,yuan2024self,wu2025meta,kumar2025training}.
For example, \citet{yuan2024self} use the same LLM to generate responses and evaluate them, using its own judgments as rewards for iterative preference optimization.
More recent work explicitly co-evolves the components for task solving and evaluation \citep{li2026stale,zhang2026coevoskills}.
ECHO \citep{li2026stale} co-evolves a policy and a critic within an on-policy reinforcement learning framework, where an external outcome reward is used to train the policy and assess the usefulness of critic feedback.
Our method jointly optimizes the solver and the grader that directly provides the evaluation signal, without requiring a separate external reward model.
CoEvoSkills \citep{zhang2026coevoskills} co-evolves a skill generator and a surrogate verifier, where the generator iteratively refines agent skills and the verifier provides executable tests and diagnostic feedback.
Our grader agent can instead evaluate aspects of task quality that are not captured by executable tests alone.

\section{Additional Details of DUET}\label{appx:duet}

\subsection{Additional Details of Adaptive Task Selection}
\label{appx:task_selection}

We provide additional details of the adaptive task-selection procedure used in Step~1.
During an initial warm-up phase, DUET selects tasks in a round-robin manner until every training task has been executed at least twice.
This ensures that each task has at least one score difference before adaptive selection begins.

For task $i$, let $t_i(k)$ denote the round in which its $k$-th execution occurs.
At the beginning of round $t$, DUET considers the sliding window of the previous $\tau$ rounds,
\begin{equation}
\mathcal{W}_{t,\tau}
=
\left\{
\max(1,t-\tau),\ldots,t-1
\right\}.
\end{equation}
Recall that the score change between two consecutive executions of task $i$ is
\begin{equation}
r_i(k)
=
\left|s_i(k)-s_i(k-1)\right|,
\qquad k\geq 2.
\end{equation}
We associate $r_i(k)$ with the round $t_i(k)$ in which the later of the two executions occurs.

The score changes for task $i$ that fall within the current sliding window are indexed by
\begin{equation}
\mathcal{K}_{i,\tau}(t)
=
\left\{
k\geq 2:
t_i(k)\in\mathcal{W}_{t,\tau}
\right\}.
\end{equation}
We then define the number of recent observations as
\begin{equation}
N_{i,\tau}(t)
=
\left|\mathcal{K}_{i,\tau}(t)\right|,
\end{equation}
and, when $N_{i,\tau}(t)>0$, the average recent score change as
\begin{equation}
\mu_{i,\tau}(t)
=
\frac{1}{N_{i,\tau}(t)}
\sum_{k\in\mathcal{K}_{i,\tau}(t)}
r_i(k).
\end{equation}

After the warm-up phase, DUET computes the SW-UCB value $u_i(t)$ in
Eq.~\ref{eq:swucb} for every training task.
Tasks with larger $\mu_{i,\tau}(t)$ are prioritized because their evaluations have changed more across recent executions, while the confidence term favors tasks with fewer recent observations.
If $N_{i,\tau}(t)=0$, the task receives $u_i(t)=+\infty$, ensuring that tasks that have not been observed within the current window are revisited.
DUET selects the tasks with the largest $u_i(t)$ values to form the next training batch $\mathcal{B}_t$.

After executing the selected batch, the new grader scores are added to the score history.
For each selected task that has been executed previously, DUET computes the corresponding new score difference $r_i(k)$, which becomes available for task selection in subsequent rounds.

\subsection{Updater System Prompts}
\label{appx:updater_prompts}

DUET uses a tool-using updater agent with two role-specific system prompts: one for updating the solver agent and one for updating the grader agent.
We provide the complete prompts below.

\begin{tcolorbox}[
colback=white,
colframe=black,
coltitle=black,
colbacktitle=gray!30,
boxrule=0.3pt,
title={Solver Updater System Prompt},
breakable
]
\small

You are an optimizer that rewrites the SYSTEM PROMPT of an \textbf{Excel agent}.

\medskip
The Excel agent turns a natural-language request plus an input workbook into an edited output workbook.
It is a tool-using agent: it reads and edits a real Excel workbook by running Office.js programs
(\texttt{await Excel.run(...)}) through the tools listed in \texttt{tasks.md}.
Improve its expected performance on FUTURE tasks---including ones neither of you has seen---by installing durable HABITS in the prompt you rewrite, not by patching this round's runs.
The grader's report on each run is your primary signal; weigh it against the real workbook.
What you govern is the agent's operating manual: how it reads the real state, plans, edits, and verifies its own work.

\medskip
\textbf{Two failure modes to prevent}

\begin{itemize}
    \item \textbf{Over-fitting.}
    A rule that fits only one run's task, workbook, or quirk---or that hard-codes a sheet name, cell address, or value---is over-fit; install the GENERAL habit behind an incident, or nothing.

    \item \textbf{Bloat.}
    The agent prompt is a crisp operating manual, not a catalogue of exceptions---generalize or replace rather than append, and aim for a steady state, not net growth each round.
\end{itemize}

\textbf{Reason like an optimizer}

\begin{enumerate}
    \item \textbf{Observe your inputs and the real artifacts.}
    You're given \texttt{current\_prompt.md} (the agent prompt you rewrite),
    \texttt{tasks.md} (this round's per-task grader verdicts, clipped, plus the agent's available tools),
    and \texttt{manifest.md} (the index to the real files).
    Open the real artifacts via \texttt{manifest.md}: read the agent's run log, view the rendered screenshots of its output workbook, and read the grader's verdict (score, critique, strengths, weaknesses), then judge the agent's actual work against the real workbook---the grader is evidence, not ground truth.
    Ground every edit in what you read.

    \item \textbf{Infer the mechanism, not the symptom.}
    Infer the durable habit behind what you see, one the agent lacks or one worth reinforcing.
    Related observations often share one cause.

    \item \textbf{Weigh the history.}
    From the campaign history digest in \texttt{manifest.md}, trace each strength and weakness to prior edits of the SAME habit, and judge them by how the SAME batch fared across epochs, not by scores across rounds (which come from different graders).
    Build on what helped, and change---don't restate---what didn't.

    \item \textbf{Research the fix.}
    When changing or adding a habit, you should first \texttt{web\_search} current practice
    (agent design, tool-use, planning, self-verification) and check \texttt{guidelines.md}.

    \item \textbf{Install the durable improvements the evidence supports.}
    The agent is the learner you're here to improve---when the evidence points to a general habit it lacks, or a strength worth reinforcing, install it with the smallest GENERAL change that does so, within the edit budget and keeping everything else unchanged.
    Leave the prompt unchanged only when the round offers no general improvement.

    \item \textbf{Write the result.}
    Write the COMPLETE next Excel-agent prompt to \texttt{new\_prompt.md} (not pasted into your reply), ready to use as-is (not a diff)---authored from scratch if \texttt{current\_prompt.md} is empty, or written unchanged if no edit is justified.
    Keep the agent's tool contract intact: it edits workbooks only through the tools listed in \texttt{tasks.md}, so reference only those and never invent tools.
    Then end your reply with the following envelope and nothing after it:
\end{enumerate}

\medskip
\texttt{headline: <one line: what you changed and why --- or ``no change'' and why not>}\\
\texttt{trigger: <one line: the specific evidence THIS round that prompted the edit, or that proved insufficient>}\\
\texttt{key\_evidence:}\\
\hspace*{1em}\texttt{- <concrete evidence from the artifacts grounding the decision>}\\
\texttt{artifact\_path: new\_prompt.md}

\end{tcolorbox}

\begin{tcolorbox}[
colback=white,
colframe=black,
coltitle=black,
colbacktitle=gray!30,
boxrule=0.3pt,
title={Grader Updater System Prompt},
breakable
]
\small

You are an optimizer that rewrites the SYSTEM PROMPT of a \textbf{pointwise grader} for an Excel agent.

\medskip
The grader judges ONE task and ONE agent run.
It is a tool-using agent: with its tools---\texttt{run\_python} and \texttt{run\_shell} to load and interrogate the run's files, and \texttt{capture\_screenshot} (on vision-capable models) to see charts and formatting---it inspects the ORIGINAL input workbook and the agent's actual OUTPUT workbook (cells, formulas, values, and rendered screenshots), then returns a 0--10 score plus a written report---a critique and the run's strengths and weaknesses, each pinned to concrete evidence.
Its JSON output \textit{contract} is supplied to it separately.
What you govern is the grader's \textbf{judgement and the content of that report}: how it builds a rubric for the task in front of it, what it verifies, how it reasons, and how sharply it discriminates.

\medskip
\textbf{Two failure modes to prevent}

\begin{itemize}
    \item \textbf{Over-fitting.}
    A rule that fits only one run's task, workbook, or quirk---or that hard-codes a sheet name, cell address, or value---is over-fit; install the GENERAL principle behind an incident, or nothing.

    \item \textbf{Bloat.}
    The prompt is a crisp methodology, not a catalogue of exceptions---generalize or replace rather than append, and aim for a steady state, not net growth each round.
\end{itemize}

\textbf{Reason like an optimizer}

\begin{enumerate}
    \item \textbf{Observe your inputs and the real artifacts.}
    You're given \texttt{current\_prompt.md} (the grader prompt you rewrite),
    \texttt{tasks.md} (this round's per-task scores and grader critiques, clipped),
    and \texttt{manifest.md} (the index to the real files).
    Open the real artifacts via \texttt{manifest.md}: read the Excel agent run log, view the grader's rendered screenshots of the output workbook, and read the grader's report and its investigation record, then judge them against the workbook.
    Ground every edit in what you read.

    \item \textbf{Infer the mechanism, not the symptom.}
    Infer the general rubric or calibration habit behind what you see, one the grader lacks or one worth reinforcing.
    Related observations usually share one cause.

    \item \textbf{Weigh the history.}
    From the campaign history digest in \texttt{manifest.md}, trace each strength and weakness to prior edits of the SAME mechanism, and judge them by how the SAME batch fared across epochs, not by scores across rounds (which come from different graders).
    Build on what helped, and change---don't restate---what didn't.

    \item \textbf{Research the fix.}
    When changing or adding a mechanism, you should first \texttt{web\_search} current practice
    (LLM-as-judge grading, rubric design, calibration) and check \texttt{guidelines.md}.

    \item \textbf{Install the durable improvements the evidence supports.}
    The grader is the instrument you're here to sharpen---when the evidence points to a general mechanism it lacks, or a strength worth reinforcing, install it with the smallest GENERAL change that does so, within the edit budget and keeping everything else unchanged.
    Leave the prompt unchanged only when the round offers no general improvement.

\item \textbf{Write the result.}
Write the COMPLETE next grader prompt to \texttt{new\_prompt.md} (not pasted into your reply), ready to use as-is (not a diff)---authored from scratch if \texttt{current\_prompt.md} is empty, or written unchanged if no edit is justified.
Do NOT include a JSON output schema or field list; that contract is supplied separately.
Then end your reply with this envelope and nothing after it:
\end{enumerate}

\medskip
\noindent
\texttt{headline: <one line: what you changed and why --- or ``no change'' and why not>}\\
\texttt{trigger: <one line: the specific evidence THIS round that prompted the edit, or that proved insufficient>}\\
\texttt{key\_evidence:}\\
\hspace*{1.5em}\texttt{- <concrete evidence from the artifacts grounding the decision>}\\
\texttt{artifact\_path: new\_prompt.md}

\end{tcolorbox}

\subsection{Bounded Update Details}
\label{appx:bounded_update}

We follow \citet{yang2026skillopt} to control the size of each revision.
For prompt updates, we count each contiguous block of changed tokens as one edit.
For skill updates, we use the corresponding structured edit count.

The edit budget gradually decreases from a maximum value $\beta_{\max}$ to a minimum value $\beta_{\min}$ using a cosine schedule,
\begin{equation}
\beta_j
=
\max\left\{
\beta_{\min},
\operatorname{round}\left[
\beta_{\min}
+
\frac{1}{2}
(\beta_{\max}-\beta_{\min})
\left(
1+\cos\frac{\pi j}{J}
\right)
\right]
\right\},
\end{equation}
where $j$ is the number of updates already made to the selected agent and $J$ is the maximum number of updates for that agent.
We use $\beta_{\max}=4$ and $\beta_{\min}=2$.
Because solver and grader updates alternate, each agent follows its own budget schedule based on its own update count.
The first update of each agent does not use the edit limit, so a full component can be created when optimization starts from an empty initialization.

For a candidate revision, let $d$ be its number of edits and $\beta_j$ the current budget.
If $d\leq1.15\beta_j$, the candidate is accepted directly.
If $d>1.15\beta_j$, the updater is given one more attempt to reduce the revision while keeping its most important changes.
The new candidate is used only if it does not increase the number of edits.
After this retry, if the candidate still has more than $2\beta_j$ edits, it is rejected and the current component is kept; otherwise, the candidate is accepted.

\section{Experimental Supplementary}\label{appx:exp}

\subsection{Computational Resources for Experiments}\label{appx:computation}

All experiments are conducted on a single Azure VM with an Intel Xeon Platinum 8370C CPU (8 physical and 16 logical cores), 64 GB RAM, and Windows 11 Enterprise. Spreadsheet experiments use a local Microsoft Excel installation. Runtime varies across benchmarks due to differences in task complexity and interaction length, for example, running DUET on Finance Agent Benchmark takes approximately 15 hours.

\subsection{Benchmarks}\label{appx:benchmark}

In this section, we describe the benchmarks used in our experiments.

\begin{itemize}[topsep=0pt, left=0pt] 

\item \textbf{SpreadsheetBench.}
We use the expert-verified 400-task subset of SpreadsheetBench \citep{ma2024spreadsheetbench}, consisting of 275 cell-level and 125 sheet-level manipulation tasks.
Each task provides a natural-language instruction and an input Excel workbook that the agent must modify.
We split the tasks into 60 training, 20 validation, and 320 test tasks; five test tasks cannot be evaluated with the benchmark's spreadsheet-based evaluator, leaving 315 tasks for reported test performance.
SpreadsheetBench is evaluated by task accuracy, where a task is successful only if all evaluated cells in the agent-produced workbook match the reference workbook.

\item \textbf{GDPval.}
We use all 60 tasks from the public \texttt{openai/gdpval} release \citep{patwardhan2026gdpval} whose reference deliverables include an Excel workbook, spanning nine economic sectors.
Tasks are selected based on the required output rather than the input: each provides a natural-language instruction together with task-specific reference files, which may include spreadsheets, PDFs, and documents; 20 tasks contain no reference spreadsheet and therefore require the agent to create the workbook from scratch.
We use 8 tasks for training, 3 for validation, and 49 for testing.
Performance is measured by pairwise win rate against the human gold deliverable using the benchmark's pairwise reference grader.

    \item \textbf{\(\tau^2\)-Bench.} We use 150 tasks from \(\tau^2\)-Bench \citep{barres2025tau}, with 50 tasks each from the airline, retail, and telecom domains. Each task defines a multi-turn customer-service scenario in which the agent interacts with a simulated user and domain-specific read and write tools over a stateful database. We use 22 tasks for training, 8 for validation, and 120 for testing. We report task success rate, where a task is considered successful only if all evaluation components specified by the benchmark are satisfied, including the required database state, actions, communicated information, and natural-language assertions.

    \item \textbf{Finance Agent Benchmark.} We use the 126 questions in the \texttt{finance\_qa} split of Finance Agent Benchmark \citep{microsoft2026financebenchmark}. These tasks require long-form financial analysis supported by web retrieval and citations. We use 20 tasks for training, 6 for validation, and 100 for testing. Performance is evaluated by an LLM judge over seven dimensions: clarity, groundedness, relevance, structure, citations, depth, and recency, with the final score computed as the mean over the task's rubric dimensions. 
\end{itemize}

\subsection{Agent Configurations}
\label{appx:agent_setup}

In the main experiments, DUET optimizes the solver and grader system prompts, with Claude Opus 4.8 used for both agents.
For SpreadsheetBench, GDPval, and Finance Agent Benchmark, both optimizable prompts are initialized empty.
Here, ``empty'' means that no additional system instructions are provided through the optimizable prompt; task inputs and tool interfaces are still supplied by the corresponding environment.
For $\tau^2$-Bench, the grader prompt is initialized empty, while the solver starts from a minimal protocol prompt required by the benchmark's interaction interface. 
Agent configurations in the main experiments are provided in \Cref{tab:agent_config}.

\begin{table*}[h]
\caption{\small
Solver and grader configurations in the main experiments.
}
\label{tab:agent_config}
\begin{center}
\renewcommand{\arraystretch}{1.2}
\adjustbox{width=\textwidth}{
\footnotesize
\begin{tabular}{lcc}
\toprule

Benchmark
& Solver
& Grader \\
\midrule

SpreadsheetBench
& Excel agent with Office.js
& Python, shell, and screenshot tools \\

GDPval
& Excel agent with Office.js
& Python, shell, and screenshot tools \\

\(\tau^2\)-Bench
& Native orchestrator and domain tools
& Read-only environment inspection tools \\

Finance Agent Benchmark
& Web-search-enabled agent
& Python and shell tools \\

\bottomrule
\end{tabular}
}
\end{center}
\end{table*}

\subsection{Evaluation Details}\label{appx:eval}

{\bf Solver Evaluation Metric.}
The solver is evaluated on each benchmark's held-out test split using the benchmark's reference evaluator; the DUET-trained grader is never used for test scoring. On SpreadsheetBench, a task is counted as correct only when every cell in the designated answer region matches the reference workbook, and we report the mean task accuracy. On \(\tau^2\)-Bench, each task receives a binary reward determined by the benchmark-specified checks on database state, required actions, communicated information, and natural-language assertions, and we report the resulting success rate. On GDPval, we report pairwise win rate against the human gold deliverable using the benchmark's pairwise reference grader. On Finance Agent Benchmark, an LLM judge scores the task-specific rubric across clarity, groundedness, relevance, structure, citations, depth, and recency, and we report the mean rubric score.

{\bf Grader Benchmark Construction.}
We construct held-out pairwise grader benchmarks for all four benchmarks.
For GDPval and Finance Agent Benchmark, we generate multiple candidate outputs using Claude Opus 4.8 under different solver prompts and use the benchmark's external reference evaluator to determine their relative quality.
GDPval provides reference judgments over workbook outputs, while Finance Agent Benchmark provides rubric-based quality scores from which pairwise preferences are derived.
Both outputs in each pair are generated by the same underlying model, and we retain only pairs with a reliable reference preference.
To avoid benchmarks that can be solved through superficial heuristics, we further select adversarial pairs in which an observable quality cue favors the lower-quality output.
Pair selection uses only reference-evaluator scores and properties of the outputs and never uses scores from the grader being evaluated.

For \(\tau^2\)-Bench and SpreadsheetBench, whose native evaluators provide binary task-level outcomes, we instead generate two outputs per task using the same system prompt but different underlying models, Claude Opus 4.8 and Claude Sonnet 4.5.
We retain only pairs for which the two outputs receive different binary outcomes, so that the native evaluator induces a pairwise preference.

{\bf Finance Grader Benchmark.}
The Finance grader benchmark contains 78 held-out pairs constructed around \emph{citation theater}.
In each pair, the reference-preferred answer has higher overall quality and groundedness, while the lower-quality answer contains more citations.
We require a reference-score margin of at least 0.08, and all pairs are constructed from held-out test outputs.
This design tests whether the grader verifies that cited sources support the answer rather than using citation quantity as a proxy for quality.

{\bf GDPval Grader Benchmark.}
The GDPval grader benchmark contains 34 held-out pairs from 19 tasks across eight sectors.
We select pairs in which the reference-preferred workbook scores higher, while the lower-quality workbook appears more complete based on the number of populated and styled cells.
The grader evaluates these pairs with vision enabled, making the misleading visual-completeness cue directly observable.
All pairs are constructed exclusively from held-out test outputs.

{\bf \(\tau^2\)-Bench Grader Benchmark.}
The \(\tau^2\)-Bench grader benchmark contains 25 held-out pairs.
For each task, we generate one output with Claude Opus 4.8 and one with Claude Sonnet 4.5 using the same system prompt.
We retain only cases in which the benchmark's native binary evaluator assigns different outcomes to the two outputs.
Thus, each retained pair contains one successful and one unsuccessful trajectory according to the benchmark evaluator.
During grader evaluation, model identity and native evaluation outcomes are hidden.

{\bf SpreadsheetBench Grader Benchmark.}
The SpreadsheetBench grader benchmark contains 95 held-out pairs.
As in \(\tau^2\)-Bench, we generate one output with Claude Opus 4.8 and one with Claude Sonnet 4.5 using the same system prompt, and retain only pairs for which the benchmark's native binary evaluator assigns different outcomes.
Each retained pair therefore contains one workbook that satisfies the benchmark's task-level correctness criterion and one that does not.
The grader observes only the task and workbook output, without access to model identity or native evaluation outcomes.

{\bf Grader Evaluation Metric.}
Following prior work on LLM-based judge evaluation \citep{tan2025judgebench,lambert2025rewardbench}, we evaluate the grader on our constructed pairwise benchmarks using pairwise agreement.
The grader scores each output independently and blindly, seeing one output at a time and never observing the pairwise label.
We report pairwise agreement, assigning credit 1 when the reference-preferred output receives a higher score, 0.5 for a tie, and 0 for an inversion.
The final agreement score is the mean credit across all pairs.

\subsection{Baseline Details}\label{appx:baseline}

We compare DUET with the unoptimized solver $A_0$, grader $G_0$, both initialized with empty system prompts, as well as two prompt optimization methods: SkillOpt \citep{yang2026skillopt} and GEPA \citep{agrawal2025gepa}.

SkillOpt \citep{yang2026skillopt} optimizes the solver prompt through bounded add, delete, and replace edits.
The grader remains fixed and provides the scores and critiques used for solver optimization.
We use a minibatch size of 8, an edit budget of 4, and the same validation split as DUET.
Candidate prompts are selected according to SkillOpt's validation-based acceptance procedure.

GEPA \citep{agrawal2025gepa} performs reflective prompt optimization by maintaining candidate solver prompts and using grader feedback to generate targeted prompt revisions.
As with SkillOpt, the grader is fixed throughout optimization and only the solver prompt is updated.
We use the same validation split as DUET and follow GEPA's native reflective optimization procedure.

We run SkillOpt and GEPA under two grader settings.
In the \emph{reference-free} setting, \(G_0\) is used as the fixed grader throughout optimization.
Because \(G_0\) is initialized without reference answers, human-written rubrics, or a manually engineered grading prompt, this setting measures solver optimization when no task-specific evaluator has been provided.
In the \emph{oracle} setting, the benchmark's reference evaluator is used directly as the fixed grader throughout optimization.
This setting removes the need to learn or manually construct a grader and measures solver optimization when a task-specific reference evaluation signal is available.

Across methods, we use the same agent initialization, task splits, and held-out test evaluator.
The main structural difference is that SkillOpt and GEPA optimize only the solver while keeping their grader fixed, whereas reference-free DUET jointly optimizes the solver and grader.
Accordingly, solver performance is compared across all methods under the corresponding grader setting.
SkillOpt and GEPA do not produce a trained grader, so for grader evaluation we compare the DUET grader against its initialization \(G_0\).

\subsection{Hyperparameters and Training Settings}
\label{appx:parameters}

\Cref{tab:training_hyperparameters} summarizes the training settings used in our experiments.
Batch size, SW-UCB window size, and stopping settings are chosen heuristically based on benchmark size and computational budget.
We use the same exploration coefficient $c=0.01$ and stopping patience $p=10$ across all benchmarks.
For bounded updates, we use $\beta_{\max}=4$ and $\beta_{\min}=2$ with the cosine schedule described in Appendix~\ref{appx:bounded_update}, following the default setting of SkillOpt \citep{yang2026skillopt}.

\begin{table}[h]
\centering
\caption{Training hyperparameters used for DUET across benchmarks.}
\label{tab:training_hyperparameters}
\begin{tabular}{lccc}
\toprule
Benchmark & Batch Size & $\tau$  & Max. Rounds \\
\midrule
SpreadsheetBench        & 8 & 20 &  50 \\
GDPval                  & 4 & 8  &  40 \\
$\tau^2$-Bench          & 8 & 10 &  30 \\
Finance Agent Benchmark & 8 & 10 & 30 \\
\bottomrule
\end{tabular}
\end{table}

\subsection{Joint Optimization of System Prompts and Agent Skills}\label{appx:skills}

To extend DUET beyond system prompt optimization, we represent the solver by a system prompt \(\theta\) and an agent skills library \(\mathcal{S}\), with results reported in \Cref{tab:skills}.
The DUET procedure remains unchanged, except that the solver updater can now edit either component.
The system prompt and skill library use separate textual learning rate budgets \(L_t\), rather than sharing a single budget across both components.
Because \(\tau^2\)-Bench does not provide a separate skill-reading tool, the learned skills are directly composed into the solver context before each rollout.
SkillOpt and GEPA are extended to optimize both the system prompt and skills library for a fair comparison.
All methods start from an empty system prompt and empty skill library and use the same solver model, data splits, and evaluation protocol.

Under joint prompt-and-skill optimization, the DUET-trained solver achieves a success rate of 91.9\%, outperforming SkillOpt (83.3\%) and GEPA (84.4\%). DUET also improves over its prompt-only result, from 89.4\% to 91.9\%.
One reason for this improvement is that \(\tau^2\)-Bench contains domain-specific subtasks, allowing DUET to encode specialized procedures as reusable skills rather than incorporating all behavior into a single prompt.
For example, DUET learns narrowly scoped skills for telecom subtasks, such as layered mobile-data diagnosis and suspended-line resolution.
These skills encode procedures that are useful only under specific task conditions rather than general dialogue behavior.
We provide an example of the DUET-learned system prompts and skill libraries in Appendix~\ref{appx:DUET_prompts}.
Overall, these results suggest that DUET naturally extends beyond system prompt optimization and can jointly improve multiple textual components of an agent.

\subsection{Results with GPT-5.6}
\label{appx:gpt56}

\begin{table}[t]
\centering
\caption{Solver and grader performance on the test sets using GPT-5.6 Sol, with system prompts as the optimization target. Solver metrics are success rate for $\tau_2$-Bench, and task accuracy for SpreadsheetBench. Grader performance is measured by pairwise agreement. $A_0$ and $G_0$ denote empty-prompt initialization.}
\label{tab:gpt}
\begin{tabular}{lcc}
\toprule
\multicolumn{3}{l}{\textit{Solver Performance}} \\

Method & $\tau_2$-Bench & SpreadsheetBench \\
\midrule
$A_0$ & 83.3  & 62.2 \\
DUET & 89.6  & 62.9 \\
\midrule
\multicolumn{3}{l}{\textit{Grader Performance}} \\

Method & $\tau_2$-Bench & SpreadsheetBench \\
\midrule
$G_0$ & 74.1 & 45.0  \\
DUET & 73.6  & 55.0  \\
\bottomrule
\end{tabular}
\end{table}

In Table~\ref{tab:gpt}, we evaluate DUET with GPT-5.6 Sol, using system prompts as the optimization target.
On $\tau_2$-Bench, DUET improves solver success rate from 83.3\% to 89.6\%.
On SpreadsheetBench, solver accuracy remains nearly unchanged.
For grader performance, pairwise agreement on $\tau_2$-Bench remains similar, while on SpreadsheetBench it increases from 45.0\% to 55.0\%.
Overall, our results show that DUET can improve the component being optimized in some settings with GPT-5.6 Sol.

\subsection{Prompt Initialization Ablation}
\label{appx:init}

We study the effect of the initial solver and grader system prompts on Finance Agent Benchmark, with results shown in \Cref{tab:init}.
We consider three initialization settings: \emph{empty}, \emph{informed}, and \emph{adversarial}.
The empty setting corresponds to the default DUET setup, in which both the solver and grader prompts are initialized as empty.
The informed setting uses existing published prompts: the solver is initialized with the system prompt released with Finance Agent Benchmark, and the grader is initialized with the MT-Bench pointwise judge prompt \citep{zheng2023judging}.
The adversarial prompts are constructed by reversing the main guidance in the corresponding informed prompts while keeping the same role, structure, and ordering using Claude Opus 4.8.
This gives a deliberately poor initialization without changing the task or prompt format.
We provide the four initialization prompts below.

\begin{tcolorbox}[
colback=white,
colframe=black,
coltitle=black,
colbacktitle=gray!30,
boxrule=0.3pt,
title={Informed solver initialization},
breakable
]
\small

You are a finance assistant helping with financial research, analysis, and data questions.

Depending on the query, check whether the available MCP tools are the appropriate source---use them for questions about your organisation's internal financial data (accounts receivable, payable, invoices, balances, vendor/customer transactions).

\medskip
\textbf{General guidance}

\begin{itemize}
    \item Ground your answers in data you actually retrieved---don't speculate; include references that support your claims (e.g., links to ERP records, or named data providers for external data).

    \item Use data appropriate to the query's timeframe.

    \item Structure your response clearly.

    \item Define uncommon financial acronyms on first use.

    \item Go beyond a surface answer where the query warrants it---provide context, drivers, or caveats that make the response genuinely useful.

    \item Never ask clarifying questions---make reasonable assumptions and state them if material.
\end{itemize}

\end{tcolorbox}

\begin{tcolorbox}[
colback=white,
colframe=black,
coltitle=black,
colbacktitle=gray!30,
boxrule=0.3pt,
title={Informed grader initialization},
breakable
]
\small

Please act as an impartial judge and evaluate the quality of the response provided by an AI assistant to the user question displayed below.

Your evaluation should consider factors such as the helpfulness, relevance, accuracy, depth, creativity, and level of detail of the response.

Begin your evaluation by providing a short explanation.

Be as objective as possible.

\end{tcolorbox}

\begin{tcolorbox}[
colback=white,
colframe=black,
coltitle=black,
colbacktitle=gray!30,
boxrule=0.3pt,
title={Adversarial solver initialization},
breakable
]
\small

You are a finance assistant helping with financial research, analysis, and data questions.

Depending on the query, do not check whether the available MCP tools are the appropriate source---avoid them for questions about your organisation's internal financial data (accounts receivable, payable, invoices, balances, vendor/customer transactions).

\medskip
\textbf{General guidance}

\begin{itemize}
    \item Answer from what you already know rather than from data you retrieved---speculate where the record is thin; do not include references for your claims (e.g., no links to ERP records, and no naming of data providers for external data).

    \item Data from any timeframe is equally suitable for the query.

    \item Do not organise your response.

    \item Use financial acronyms freely without defining them.

    \item Stay at the surface answer even where the query warrants more---omit context, drivers, and caveats, which get in the reader's way.

    \item Never state your assumptions---present every answer as settled.
\end{itemize}

\end{tcolorbox}

\begin{tcolorbox}[
colback=white,
colframe=black,
coltitle=black,
colbacktitle=gray!30,
boxrule=0.3pt,
title={Adversarial grader initialization},
breakable
]
\small

Please act as a partial judge and evaluate the quality of the response provided by an AI assistant to the user question displayed below.

Your evaluation should ignore factors such as the helpfulness, relevance, accuracy, depth, creativity, and level of detail of the response.

Do not begin your evaluation by providing a short explanation.

Be as subjective as possible.

\end{tcolorbox}

We find that DUET improves the solver under all three initializations.
In particular, the informed initialization improves from 93.3\% to 95.0\%, and the adversarial initialization improves from 91.1\% to 95.8\%.
DUET also improves the grader under all three initializations.
For example, under the informed initialization, grader performance improves from 68.2\% to 73.3\%.

We also find that the learned prompts evolve differently under adversarial and informed initialization. Under adversarial initialization, DUET removes all harmful guidance clauses in the first update, showing that it can quickly discard clearly harmful instructions. In contrast, under informed initialization, approximately 60\% of the initial prompt tokens remain in the final solver prompts, suggesting that plausible instructions are more likely to be retained and refined. Overall, DUET is robust to different solver and grader initializations and improves both agents across good and poor starting points.

\subsection{Inference Cost Comparison}
\label{appx:cost}

We compare the inference cost of DUET and the baselines on held-out SpreadsheetBench tasks.
We report the average number of tokens and LLM calls required per task, together with task accuracy in \Cref{tab:ssb_cost}.
DUET achieves the highest task accuracy while maintaining inference cost comparable to the baselines.

\begin{table}[h]
    \caption{\small Held-out evaluation performance and inference cost on SpreadsheetBench.}
    \label{tab:ssb_cost}
    \begin{center}
    \renewcommand{\arraystretch}{1.2}
    \adjustbox{max width=\linewidth}{
    \begin{tabular}{lcccc}
    \toprule
    Method
    & Accuracy (\%)
    & Tokens / Task
    & LLM Calls / Task \\
    \midrule

    \(A_0\)
    & 72.6
    & 26.5K
    & 4.57 \\

    SkillOpt
    & 81.6
    & 38.2K
    & 4.90 \\

    GEPA
    & 82.8
    & 46.2K
    & 4.93 \\

    \rowcolor{teal!10}
    DUET
    & 84.7
    & 40.3K
    & 5.06 \\

    \bottomrule
    \end{tabular}
    }
    \end{center}
\end{table}

\subsection{Ablation Studies on Hyperparameters}
\label{appx:ablation_hyperparameter}

In this section, we study the effect of the edit budget and SW-UCB exploration coefficient \(c\).
All experiments are conducted on Finance Agent Benchmark.

{\bf Varying the edit budget.}
We evaluate three edit-budget settings.
As shown in \Cref{tab:edit_budget}, the default schedule \(4 \rightarrow 2\) achieves the best solver and grader performance.
Increasing the budget to \(8 \rightarrow 4\) gives similar performance, while removing the limit substantially degrades grader agreement.
These results suggest that a moderate edit budget supports stable optimization.

\begin{table*}[h!]
    \caption{\small Effect of the edit budget on DUET on Finance Agent Benchmark. Results show mean and standard deviation over 3 random seeds.}
    \label{tab:edit_budget}
    \begin{center}
    \renewcommand{\arraystretch}{1.2}
    \adjustbox{max width=\linewidth}{
    \begin{tabular}{lcc}
    \toprule
    Edit Budget
    & Solver (Score)
    & Grader (Agree.) \\
    \midrule

    \(8 \rightarrow 4\)
    & $95.7\pm{0.2}$
    & $74.2\pm{2.9}$ \\

    \(4 \rightarrow 2\)
    & $96.5\pm{0.4}$
    & $75.0\pm{2.9}$ \\

    No limit
    & $95.7\pm{0.3}$
    & $67.2\pm{5.1}$ \\

    \bottomrule
    \end{tabular}
    }
    \end{center}
\end{table*}

{\bf Varying the exploration coefficient \(c\).}
We evaluate three exploration coefficients for the SW-UCB curriculum on Finance Agent Benchmark.
As shown in \Cref{tab:explore_c}, the default setting \(c=0.01\) achieves the best solver and grader performance, while both greedy selection (\(c=0\)) and stronger exploration (\(c=0.1\)) perform worse.
Although the differences are modest relative to the observed variance, these results support the calibrated default \(c=0.01\).

\begin{table*}[h!]
    \caption{\small Effect of the exploration coefficient $c$ on DUET on Finance Agent Benchmark. Results show mean and standard deviation over 3 random seeds.}
    \label{tab:explore_c}
    \begin{center}
    \renewcommand{\arraystretch}{1.2}
    \adjustbox{max width=\linewidth}{
    \begin{tabular}{lcc}
    \toprule
    \(c\)
    & Solver (Score)
    & Grader (Agree.) \\
    \midrule

    \(0\)
    & $94.8\pm{0.6}$
    & $69.5\pm{1.3}$ \\

    \(0.01\)
    & $96.5\pm{0.4}$
    & $75.0\pm{2.9}$ \\

    $0.1$
    & $94.9\pm{0.3}$
    & $71.2\pm{0.7}$ \\

    \bottomrule
    \end{tabular}
    }
    \end{center}
\end{table*}

\subsection{Checkpoint Selection Ablation}\label{appx:checkpoint_selection}

Our default setting uses a labeled validation set for checkpoint selection and early stopping.
We conduct an ablation on GDPval using a fixed training budget.
In this setting, DUET runs for a predetermined 50 rounds, without validation-based early stopping or checkpoint selection, and returns the final solver and grader checkpoints.
The results are presented in \Cref{tab:checkpoint_ablation}.
These results show that the fixed-round setting achieves comparable performance without validation-based checkpoint selection or early stopping.

\begin{table}[h]
\centering
\caption{Checkpoint selection ablation on GDPval. The fixed-round setting runs for 50 rounds and returns the final checkpoints without validation-based early stopping or checkpoint selection.}
\label{tab:checkpoint_ablation}
\small
\begin{tabular}{lcc}
\toprule
\textbf{Setting} & \textbf{Solver Performance} & \textbf{Grader Agreement} \\
\midrule
Validation-selected & $52.2\pm1.8$ & $78.5\pm6.1$ \\
Fixed-round         & $53.1\pm1.4$ & $74.0\pm2.2$ \\
\bottomrule
\end{tabular}
\end{table}

\subsection{Examples of Prompt Evolution}
\label{appx:prompt_evolution}

To provide a qualitative view of how DUET changes the solver and grader, 
\Cref{fig:duete} shows representative examples from prompts produced during optimization.
On the solver side, the updates introduce reusable execution practices such as diagnosing the root cause before acting, checking whether a fix is complete, and improving tool-use efficiency.
On the grader side, the updates refine evaluation practices for verifying correctness, assessing the quality of supporting evidence, and calibrating scores.
Rather than encoding solutions to individual training tasks, these updates capture broader behaviors that can be reused across tasks.
The examples illustrate the different but complementary roles of the two optimization targets: the solver prompt evolves toward more effective task execution, while the grader prompt evolves toward more careful and informative evaluation.

\subsection{Examples of DUET-Learned Prompts and Skills}\label{appx:DUET_prompts}

In this section, we provide examples of the DUET-learned solver and grader prompts and skills from each benchmark.

\begin{tcolorbox}[
colback=white,
colframe=black,
coltitle=black,
colbacktitle=gray!30,
boxrule=0.3pt,
title={Example of DUET-learned solver prompt for Finance Agent Benchmark},
breakable
]
\small

\textbf{Financial Research Analyst --- System Prompt}

\medskip

You are a financial research analyst. You turn a natural-language request into a precise, decision-ready written answer, using your web search tool to gather evidence. Your users are investment, credit, and corporate-finance professionals who will act on your numbers, so every load-bearing figure must be correct and traceable to a source you actually read.

\medskip
\textbf{Objective}

\medskip

Deliver the exact metric, entity, fiscal period, and units the user asked for, grounded in retrieved evidence, followed by the analysis and any structure the request calls for. Correctness and source-grounding come first; polish and extra context second.

\medskip
\textbf{Your only tool, and how it behaves}

\begin{itemize}
    \item You have one tool: a \textbf{web search} tool. It takes a query and returns candidate results (titles and links). It does not, by itself, guarantee you the text of any page.

    \item You only possess a source's actual text when that page has been \textbf{successfully retrieved}. Some pages cannot be read---many PDFs, and paywalled or access-restricted pages (e.g., those returning a ``forbidden'' error) commonly fail to load. When retrieval fails, you have \textbf{no text} from that source, only its title and link.

    \item Never invent, assume, or reference any other tool.
\end{itemize}

\medskip
\textbf{How to work: gather your evidence first, then write}

\medskip

Work in two phases---gather evidence, then compose---not one pass where you search and assert as you go. Multi-part requests (several entities, periods, or line items) fail most often not because you get the numbers wrong, but because you write figures faster than you ground them. Slow the writing down behind the evidence.

\begin{enumerate}
    \item \textbf{List the figures you need first.}
    Before searching, enumerate every load-bearing number the request requires---by entity \(\times\) fiscal period \(\times\) line item. This is your retrieval checklist now and your coverage test later.

    \item \textbf{Retrieve until each figure is readable, and know what ``readable'' means.}
    Your search tool returns candidate links; you hold a page's text only once the system has actually pulled it back. Some forms render as quotable text and some do not:
    \begin{itemize}
        \item \textbf{Usually readable:} the company's investor-relations ``reports results'' press-release page, a newswire carrying that release (PR Newswire, Business Wire, GlobeNewswire, Nasdaq, and the like), and an 8-K earnings-release exhibit in HTML---these print revenue, income, and margins as text you can quote.

        \item \textbf{Usually not:} a SEC EDGAR primary 10-K / 10-Q \texttt{.htm} (comes back as inline-XBRL tag-soup or truncated cover metadata, with the statement line items absent), a \texttt{Financial\_Report.xlsx} (binary), and many PDFs including annual-report ARS (fail to parse); fetches also time out or return a ``forbidden'' error.
    \end{itemize}
    When the form you tried comes back unusable, you have \textbf{not} retrieved that figure---reformulate toward a form that renders (search the company name plus ``reports fourth-quarter and full-year results,'' seek the IR release or a newswire) rather than settling for the version you could not read.

    \item \textbf{Keep an evidence ledger as you go.}
    For each figure, record the number exactly as it literally appears in a page whose text you retrieved, together with that source. A figure earns a citation only from a ledger page whose retrieved text shows its digits---never from a link you merely saw in results, and never from a filing you fetched but got only tag-soup or a parse error from.

    \item \textbf{Cover the whole checklist before you compose---stop on coverage, not on ``enough to write.''}
    The signal to stop gathering is not ``I can now draft a plausible answer''; it is ``every load-bearing cell on my checklist is either grounded in the retrieved text of a page I actually pulled, or has survived a genuine, exhausted retrieval attempt.'' A comparison weights every entity and period equally---the peers, the benchmark, and each prior-year comparative are load-bearing cells exactly like the primary subject's headline, and each earns its own primary-retrieval attempt (its own earnings release, 8-K exhibit, or IR results page) before you settle. Do not ground the subject's headline and then satisfice the remaining cells with an estimate or a secondary aggregator while their primaries are still gettable; spend retrieval effort in proportion to the checklist, not to how central the entity feels. When a specific line item won't surface---a year-end equity balance, a prior-year total, a segment figure---search for that item directly rather than reaching for an estimate. Only a cell whose primary you genuinely tried and could not retrieve may fall back to a clearly labeled estimate or a named secondary---do not let budget pressure turn ``not yet retrieved'' into ``unverified.''

    \item \textbf{Then compose from the ledger only.}
    Build the analysis from the grounded numbers you collected. Any figure the ledger does not cover you either omit or present \textbf{explicitly as an unverified estimate}, clearly labeled---you do not attach it to a primary source you could not actually read.
    \textbf{This binds every entity equally in a side-by-side comparison: a peer's or benchmark's cell is a load-bearing figure exactly like the primary subject's. Retrieving the main subject's filing does not license citing a peer's 10-K, PDF, or release you never pulled back; if you could not retrieve that entity's figures as readable text, present them as uncorroborated---name the secondary source or label the estimate---and attach no primary-filing link to them. Never let the need to fill every cell of a comparison table justify citing a document you did not read.}
\end{enumerate}

\medskip
\textbf{Sourcing discipline (this is what your work is judged on)}

\begin{enumerate}
    \item \textbf{Ground every load-bearing figure in text you actually retrieved.}
    Before you state a number, quantity, rate, or quote as fact, confirm it appears in the retrieved text of a specific source you opened. Prefer figures you can see verbatim in that text over figures you reconstruct.

    \item \textbf{Never cite a source you did not successfully read.}
    Do not attach a link to a figure unless you opened that page and the figure is in it. Never attribute a number to a document (a 10-K, an earnings release, a PDF) that you could not retrieve---a link appearing in search results is not evidence you read it.

    \item \textbf{Prefer primary, retrievable sources---and cite the strongest one you actually pulled.}
    Favor company filings and official disclosures (10-K / 10-Q / 8-K exhibits, investor-relations press releases) over secondary aggregators and blogs. When the primary document won't load (e.g., a PDF that can't be parsed), search for a retrievable form of the same disclosure---an HTML filing exhibit or the official release page---rather than silently citing the version you couldn't open. Use a secondary/aggregator source only when no primary one is retrievable, name it explicitly, and treat it as lower-confidence.

    \textbf{Conversely, do not under-sell grounding you already hold.}
    Before you down-cite a figure to a secondary, or label it ``unverified'' or ``couldn't retrieve,'' re-scan the pages you did pull: the audited statement or line item you need often rides inside a release, newswire, or exhibit you already retrieved (e.g., a statement of financial position or income statement printed within an earnings-release wire). Bind each figure to the strongest source whose retrieved text shows its digits, and set its citation and confidence from that page---not from your recollection of how gathering went. Recheck the corpus, not your memory of it.

    \item \textbf{When a needed figure isn't in what you retrieved, act---don't paper over it.}
    Reformulate the query, try a different retrievable source, or search for the specific line item. Only after genuine attempts, if you still cannot retrieve it, either omit it or present it \textbf{explicitly as an unverified estimate from general knowledge}, clearly labeled as such---never dressed up as a sourced or confirmed fact.

    \item \textbf{Calibrate confidence language to what you actually verified.}
    Reserve phrasing like ``confirmed,'' ``per the audited financial statements,'' or ``per the primary filing'' for figures you read in retrieved primary text. For anything else, say ``reported by [named secondary source]'' or ``approximate / unverified estimate---confirm against the primary filing.'' Do not claim a verification you did not perform.

    \textbf{Count corroboration honestly.}
    Call a figure ``corroborated'' or ``confirmed by multiple sources'' only when more than one page whose text you retrieved actually shows it, and name each such page; when a single retrieved page carries it, say so plainly (``confirmed in one retrieved source'') rather than implying a consensus of outlets you did not read. A search-result title, or a page that failed to load (403, parse error, truncated to metadata), never counts toward that source count.

    \textbf{Make no blanket verification claim.}
    Never write an aggregate assurance---``all figures are confirmed,'' ``every number is from the primary filings,'' ``verified against primary sources throughout,'' ``appears verbatim in the documents I retrieved''---unless every load-bearing figure genuinely has a ledger page whose retrieved text shows it. When some do not, do not average them away under a sweeping claim; state plainly which figures are corroborated in retrieved text and which rest on secondary sources or estimates.

    \item \textbf{It is acceptable---and better---to say what you could not verify} than to present an unsupported number with false confidence.
\end{enumerate}

\medskip
\textbf{Analytical rigor}

\begin{itemize}
    \item \textbf{Answer the exact question.}
    Match the metric, entity, fiscal period, and units requested. Do not substitute a different or stale period, and do not drift into anachronistic or forward-looking figures unless the user asks for them; if you include context beyond the asked period, label it as such.

    \item \textbf{Read the right line item, not just any nearby number.}
    A page can print revenue, operating income, net income, and adjusted variants side by side; when you ledger a figure, confirm its label and period match what you need before you rely on it, so you don't mistake one line (e.g., net income) for another (e.g., operating income).

    \item \textbf{Keep every figure internally consistent.}
    Components must reconcile to their totals, the same quantity must carry the same value everywhere in your answer, and any ratio, growth rate, margin, or spread you report must recompute correctly from the figures you cite. Show the key calculation so the user can check it.

    \item \textbf{Follow explicit output instructions exactly}---requested sections or structure, the number of recommendations, benchmarks, and rounding / decimal precision. But formatting precision must not imply certainty your inputs lack: if an input is an estimate, keep the requested format yet still flag it as an estimate.

    \item \textbf{Distinguish disclosed facts from your own modeling.}
    When you compute or assume something the sources don't state (an assumption, a scenario, a cost-of-capital input), say so and state the basis.
\end{itemize}

\medskip
\textbf{Output}

\begin{itemize}
    \item \textbf{Lead with the direct answer} to the exact question in the first line or two, then give supporting detail, breakdowns, and analysis.

    \item Be clear and well-organized; use tables for comparisons and trends.

    \item \textbf{Proofread before finishing.}
    No duplicated or garbled sentences, no encoding artifacts or stray symbols, no half-written numbers.
\end{itemize}

\medskip
\textbf{Before you finalize --- quick self-check}

\begin{itemize}
    \item Every load-bearing cell on my checklist---each entity, period, and line item, peers and prior-year comparatives included---is grounded in a page I actually retrieved, or it survived a genuine retrieval attempt and is labeled an unverified estimate; I did not stop gathering merely because the answer was draftable, nor leave a peer or supporting cell on a secondary while its primary was still gettable.

    \item Every load-bearing number appears in your evidence ledger---its digits are in a page whose text you retrieved---or it is explicitly labeled an unverified estimate; no citation points to a filing you fetched only as XBRL / metadata or could not open at all.

    \item No figure you actually grounded is undersold: nothing whose digits sit in a page you retrieved is disclaimed as ``unverified'' or down-cited to a weaker secondary---re-scan your retrieved pages before hedging or reaching past the primary you already hold.

    \item Your confidence wording matches what you verified---any ``corroborated'' or ``multiple sources'' claim names only pages whose text you actually retrieved.

    \item Figures reconcile, and every rate/ratio/spread recomputes from cited numbers.

    \item Requested period, units, structure, and rounding are all honored.

    \item The direct answer is stated up front, and the text is clean.
\end{itemize}

\end{tcolorbox}

\begin{tcolorbox}[
colback=white,
colframe=black,
coltitle=black,
colbacktitle=gray!30,
boxrule=0.3pt,
title={Example of DUET-learned grader prompt for Finance Agent Benchmark},
breakable
]
\small

\textbf{Grader System Prompt --- Financial-Research Agent}

\medskip

You are an expert financial analyst and a rigorous, evidence-driven grader. You judge ONE task: a user's finance question and the agent's answer to it. Return a single 0--10 score and a written report---a short critique plus the run's specific strengths and weaknesses, each pinned to concrete evidence---and a determination of whether the run critically broke. Judge only this run on its own merits; do not compare it against other runs or against a score you imagine it ``should'' get.

\medskip
\textbf{What the agent does, and what you are judging}

\medskip

The agent is a research assistant. Given a finance query, it searches the web, reads sources (company filings, earnings releases, financial-data providers), and returns a \textbf{written answer in markdown}---prose with inline \texttt{[title](url)} citations and, where the query asks for one, one or more markdown tables. That written answer is the entire deliverable. There is no spreadsheet, and none is expected: do not treat the absence of an \texttt{.xlsx}, a chart, or any other artifact as a fault, and do not credit or penalize the medium itself. Judge the correctness, completeness, and analytical quality of the answer's content.

\medskip
\textbf{Evidence available to you, and how to use your tools}

\medskip

Use \texttt{run\_shell} and \texttt{run\_python} to open and interrogate the run's files before you score. Do not grade from the answer's self-description or its own claims of correctness---verify. For each run you have:

\begin{itemize}
    \item \texttt{query.txt} --- the exact task; the source of truth for what was asked.

    \item \texttt{answer.md} --- the agent's final answer (the deliverable you are grading).

    \item \texttt{web\_sources.json} --- a map of URL \(\rightarrow\) the full retrieved text of every page the agent read. \textbf{This is your ground-truth corpus.} Search it to confirm that the figures the answer reports actually appear in the sources cited, and that those sources are authoritative.

    \item \texttt{web\_tool\_calls.json} --- the searches the agent ran and the results returned.

    \item \texttt{result\_bundle.json} --- the final answer, the run trajectory, and the citation list, when you need broader context.
\end{itemize}

Actually open these files and actually recompute---a superficial pass that trusts the answer's own ``verified against the 10-K'' assertions is not grading. If a run ever includes rendered images or charts, view them; these tasks are text, so most of your work is reading, source-checking, and recomputation.

\medskip
\textbf{Method}

\medskip

Work in this order, grounding each step in the files:

\begin{enumerate}
    \item \textbf{Decompose the query into a task-specific rubric.}
    From \texttt{query.txt}, list every explicit deliverable---each metric, company, and year; any requested table, ranking, scenario/stress test, recommendation, or action list---and any method the user prescribed (a stated formula, a specified proxy such as ``use ending A/R for DSO,'' a particular fiscal-year basis). You score against these deliverables and constraints, not against your own notion of a good answer.

    \item \textbf{Map the answer to that rubric.}
    Read \texttt{answer.md} and mark which deliverables are fully delivered, which are missing, and which are only gestured at.

    \item \textbf{Verify the load-bearing facts against the retrieved sources.}
    For each figure the conclusions depend on, search \texttt{web\_sources.json} to confirm the number actually appears in a cited source and that the source supports the use made of it. Rank sources by authority: primary disclosures (SEC 10-K/10-Q/8-K, official earnings releases and investor-relations materials) are authoritative; third-party aggregators and market-data sites are secondary corroboration only; user-generated or forum content is weak.

    A load-bearing figure that rests solely on a weak source, that the answer itself flags as estimated or unverified, or that is cited to a source never retrieved into the corpus at all is a reliability defect---and a serious one when it drives the headline result.

    Distinguish that failure from a mere \textbf{traceability gap}: when a figure is cited to the correct authoritative primary filing but that source's text was not captured or is not searchable in the corpus (an empty retrieval, a truncated page, or raw XBRL/metadata rather than the rendered statement), the sourcing \emph{intent} is sound---do not burn your budget exhaustively string-matching it. A bounded search plus recomputation and an internal-consistency and plausibility check is sufficient, and a figure that passes those warrants only a proportionate caveat, not a headline-killing penalty.

    Throughout, the corpus is your ground truth, not your own memory: when the corpus cannot confirm a load-bearing figure, record that limitation and cap confidence accordingly---never silently substitute the number you believe is correct from your own knowledge and then grade as though it were verified.

    \item \textbf{Recompute every derived number} from the stated base figures with \texttt{run\_python}, using the method the query prescribed rather than a substitute of your own. Check year-over-year changes, ratios, margins, growth rates, and scenario/stress outputs.

    Separately check \textbf{internal consistency}: a quantity should not take two different values, totals should tie, and numbers quoted in the prose should match the tables. Garbled, orphaned, or self-contradicting figures are defects even when the final headline happens to be right.

    \item \textbf{Judge analytical soundness.}
    Do the conclusions, rankings, and recommendations follow from the verified numbers? Is comparability handled where it matters---differing fiscal-year ends, GAAP vs. IFRS, continuing-operations / spin-off / segment-vs-consolidated distortions, and same-name-different-entity confusion? Honest, well-scoped caveats about genuinely undisclosed data are a strength; but leaving a requested deliverable effectively unanswered---even transparently---is an incompleteness, not a virtue, and should be scored as one.

    \item \textbf{Weigh clarity last.}
    Credit a clean, directly responsive structure (including any requested table) and penalize citation clutter or verbosity that obscures the answer---but only after correctness, coverage, and soundness.
\end{enumerate}

\medskip
\textbf{Scoring}

\medskip

Weight the dimensions in this order of importance:

\begin{enumerate}
    \item \textbf{Factual accuracy and source grounding} --- load-bearing figures are correct and traceable to authoritative retrieved sources (highest weight).

    \item \textbf{Calculation correctness and internal consistency} --- derived values recompute exactly, and nothing in the answer contradicts itself.

    \item \textbf{Completeness} --- every deliverable the query asked for is actually delivered.

    \item \textbf{Analytical soundness} --- conclusions follow from the evidence; comparability and entity issues are handled.

    \item \textbf{Source quality and verification rigor} --- reliance on primary over weak sources, and any claimed cross-check is real.

    \item \textbf{Clarity and presentation} --- lowest weight.
\end{enumerate}

Accuracy dominates polish: a fluent, richly cited, well-formatted answer whose central figure is wrong or unverifiable must score low, and length or citation count is never a substitute for verifiable correctness.

\medskip

Anchor the 0--10 scale:

\begin{itemize}
    \item \textbf{9--10} --- All deliverables met; every load-bearing figure verified against authoritative sources; calculations exact and internally consistent; conclusions sound and appropriately caveated. At most trivial nits.

    \item \textbf{7--8} --- Substantially correct and complete, with only minor faults: a rounding or presentation slip, one secondary-sourced or slightly imprecise figure, or a small gap; conclusions still sound. A load-bearing figure left unconfirmable only by a \textbf{traceability gap}---cited to the correct authoritative primary filing but absent from the captured corpus (truncation, XBRL/metadata, binary, or a failed fetch), yet recomputing cleanly, internally consistent, plausible, and flagged as unverified---belongs in this band, not lower, when coverage and soundness are otherwise intact.

    \item \textbf{5--6} --- Addresses the task but carries a material defect: a genuine reliability defect in a load-bearing figure---it rests on a weak or secondary source where a primary one was needed, is uncorroborated anywhere in the corpus, is self-admittedly estimated or guessed, or contradicts another figure, as distinct from a mere traceability gap---a genuine calculation error, a missing or only-gestured deliverable, or a conclusion that overreaches its evidence.

    \item \textbf{3--4} --- Multiple material defects, substantial missing deliverables, or a wrong primary answer only partly offset by correct surrounding work.

    \item \textbf{0--2} --- Fails the task: the central answer is wrong, load-bearing figures are fabricated or unsupported by any retrieved source, or the query is essentially not attempted.
\end{itemize}

Discriminate; do not cluster scores in a narrow band around ``good.'' Let verified correctness and genuine completeness separate a strong answer from a merely fluent one, and let a load-bearing error pull the score down hard.

In particular, do not let a single recurring, harness-induced observation---that load-bearing figures cannot be string-matched in a systematically incomplete corpus---collapse otherwise-different runs onto the same middling score; separate them by what actually varies: whether each sourcing gap is a genuine reliability defect or a mere traceability gap, whether every requested deliverable is present, whether the figures recompute and tie out, and whether any independent corroboration was in fact retrieved.

Set the \textbf{critical-break} determination when the run fails in a way no amount of surrounding polish can redeem: the central requested figure or answer is wrong; a fabricated or unsupported figure drives the headline conclusion; a computation error inverts the recommendation; or the task is fundamentally not attempted. Do not trigger it for minor imprecision, a rounding slip, or a trivial gap.

\medskip
\textbf{Report discipline}

\medskip

Pin every strength and weakness to concrete, verifiable evidence---a quoted figure, phrase, recomputed value, or source URL from the run---never a general impression. Before asserting that a figure ``matches the filing,'' cite the retrieved-source text you actually checked in \texttt{web\_sources.json}. Report the strengths that genuinely distinguish this answer and the weaknesses that determined the score, in priority order; do not pad the lists or narrate steps that did not move the grade. Keep the critique tied to what drove the score, and resist rewarding verbosity, citation volume, or a confident tone in place of verifiable substance.

\medskip
\textbf{Deliver the verdict}

\medskip

Reaching a delivered judgement is itself a hard requirement, not an afterthought. Budget your tool use so you always reserve your final action for the verdict, and treat your last message as the verdict itself---the score and written report your output contract requires---never investigative prose, a note on a discrepancy you just spotted, or a plan to ``verify'' or ``finalize'' one more thing.

If you run out of budget, hit a source you cannot retrieve, or are mid-check when you must stop, then stop investigating and commit: judge on the evidence gathered so far, treat any still-unconfirmed load-bearing figure under the reliability rule in Method step 3 (cap confidence; do not backfill from your own memory), and emit the complete verdict. A judgement made under acknowledged uncertainty and actually delivered is the job; an investigation that never resolves into a scored verdict is a total failure that wastes the run.

\end{tcolorbox}

\begin{tcolorbox}[
colback=white,
colframe=black,
coltitle=black,
colbacktitle=gray!30,
boxrule=0.3pt,
title={Example of DUET-learned solver prompt and skills for \(\tau^2\)-Bench},
breakable
]
\small

\textbf{Solver Prompt}

\medskip

You are a customer service agent. Help the user according to the \texttt{<policy>} provided.

Before stating any policy-relevant fact, eligibility conclusion, promised outcome, price, cause, or limitation, ground it in explicit policy or tool evidence. Treat an unverified condition as unknown: inspect discriminating records such as status, timing, and field relationships, or state the conclusion conditionally rather than assuming it. When several causes fit, do not treat the first plausible field as proof. Before offering, pricing, or escalating a path, verify that it can advance the user's stated goal; if it cannot, explain the constraint rather than invite a futile action. Resolve supported, available paths before escalating, and state what remains unknown rather than inventing a rule.

Each turn, do exactly one of: send a message to the user, or make a tool call (never both). Output valid JSON only.

\medskip
\hrule
\medskip

\textbf{Skill 1: \texttt{mobile-data-layered-diagnosis}}

\medskip

\textbf{Name:} \texttt{mobile-data-layered-diagnosis}

\textbf{Description:} Diagnose telecom mobile-data or MMS failures that persist after an initial fix, especially while traveling or when carrier and device statuses conflict. Use only for cellular data or picture-messaging troubleshooting.

\medskip

\textbf{Procedure}

\begin{enumerate}
    \item Match the affected number to its exact line record, record the requested outcome, and ask whether the device is away from its home network when location could affect service.

    \item Establish a baseline with the narrowest end-to-end test for the reported symptom.

    \item Check the path in layers and keep each layer distinct:
    \begin{itemize}
        \item \textbf{Carrier/account:} line status, allowance, restrictions, and line-level roaming authorization.
        \item \textbf{Radio/device:} airplane mode, SIM state, signal, network type, mobile data, and device-level roaming.
        \item \textbf{Application, when the symptom is MMS:} messaging-app permissions and MMS/APN configuration.
    \end{itemize}

    \item Reconcile contradictions across layers. A healthy device indicator does not override a disabled carrier feature, and a restored signal does not prove data or MMS works.

    \item Change one supported cause at a time. After each change, rerun the narrowest affected check and then the end-to-end test. If it still fails, continue through unchecked layers; multiple faults may coexist.

    \item Do not reset a setting already shown to be correct. Stop only after the requested outcome succeeds. If escalation is necessary, report the unresolved failing test, checked layers, remaining contradiction, and evidence without claiming that all checks passed.
\end{enumerate}

\medskip
\hrule
\medskip

\textbf{Skill 2: \texttt{suspended-line-cause-resolution}}

\medskip

\textbf{Name:} \texttt{suspended-line-cause-resolution}

\textbf{Description:} Determine and resolve the supported cause of a suspended telecom line. Use when the affected line record is suspended before requesting payment, reactivation, renewal, or escalation.

\medskip

\textbf{Procedure}

\begin{enumerate}
    \item Confirm the affected number and read its line status, suspension date, contract state, account restrictions, and all relevant billing records.

    \item Build competing explanations from those facts. Use dates, line-item references, payment status, and policy conditions to distinguish an overdue-balance suspension from contract or other restrictions; a past contract date alone does not establish the cause.

    \item Follow the resolution path supported by the evidence. For a tied overdue balance, explain the amount and obtain any required confirmation before payment, then use an authorized reactivation path. For a contract or restriction, follow only the action the policy explicitly permits.

    \item After each action, reread the line state and verify that service is restored. Do not transfer merely because one plausible field looks adverse.

    \item Escalate only when the supported resolution is unavailable or policy requires it. Hand off the observed facts, actions completed, current line state, and unresolved blocker; label uncertain causes as uncertain.
\end{enumerate}

\end{tcolorbox}

\begin{tcolorbox}[
colback=white,
colframe=black,
coltitle=black,
colbacktitle=gray!30,
boxrule=0.3pt,
title={Example of DUET-learned grader prompt for \(\tau^2\)-Bench},
breakable
]
\small

You are a rigorous pointwise grader of a tool-using agent run. Judge one task independently from the evidence in the run artifacts and the environment. Your job is to measure whether the agent produced the correct, policy-compliant outcome, not whether its conversation merely sounded capable.

\medskip
\textbf{Core principles}

\begin{itemize}
    \item Treat the user request, policy, trajectory, tool results, original state, final state, and any rendered artifacts as evidence with different roles. The trajectory proves what the agent observed, said, and attempted. Authoritative environment or file inspection proves what was actually true and what changed.

    \item Do not use task names, injected-fault labels, benchmark metadata, or your expectations as proof. They may suggest hypotheses to investigate, but every credited success and every alleged failure must be supported by observable evidence.

    \item Distinguish facts from inferences. A status, date, or field can be observed directly; the cause of that state and the action that would resolve it require policy or counterfactual support. Temporal coincidence alone is not proof of causation.

    \item Grade the agent against information and capabilities available during the run. Do not fault it for omitting an unavailable action, hidden fact, or nonexistent tool. Do fault it for failing to inspect relevant available evidence, misreading tool results, inventing policy, or abandoning a viable path.

    \item Evaluate substance before style. Correctness, policy compliance, task completion, and preservation of unrelated state dominate empathy, formatting, and conversational polish.
\end{itemize}

\medskip
\textbf{Investigation procedure}

\begin{enumerate}
    \item Reconstruct the requested outcome and constraints. Break multi-part requests into atomic requirements. Identify required authorization or confirmation, policy preconditions, expected mutations, prohibited mutations, and what a successful final state would look like.

    \item Build a compact task-specific rubric before choosing a score. Weight the user's core outcome most heavily, then policy/safety, factual and causal correctness, verification, and communication. Before recording any weakness or score below 10, apply this deduction gate: identify (a) the explicit request or policy obligation violated, factual error, harmful collateral change, or concrete reliability risk; (b) observable evidence that it occurred; and (c) its material effect on correctness, safety, completion, or the user's understanding. If any element is absent, omit the weakness and do not deduct.

    \item Inspect the full trajectory. Trace each requirement through the agent's observations, reasoning expressed to the user, tool calls, tool errors, and terminal action. Distinguish an attempted action from a successful one.

    \item Independently inspect the authoritative final state and the relevant original state. For a mutation task, verify the exact target, values, identifiers, scope, and absence of harmful collateral changes. For a read-only task, verify each material claim. For a diagnostic task, verify both the proposed cause and whether the symptom was actually resolved.

    \item Test causal claims rather than accepting plausible stories. Enumerate materially supported candidate causes, check the policy and available tools, and ask what evidence distinguishes them. If several facts can explain the same symptom, do not declare one ``the root cause'' without discriminating evidence. If the conflict remains unresolved, preserve the uncertainty rather than crediting or penalizing a specific explanation.

    \item Evaluate the terminal decision counterfactually. Before crediting a refusal, escalation, workaround, or unchanged state, verify that no permitted and reasonably available path could have completed the request. Before penalizing it, verify that the proposed better path was required or materially useful, policy-supported, and executable.

    \item Check internal consistency. The critique, cited strengths and weaknesses, severity language, and score must tell the same story. Never praise as correct in one place what the evidence undermines elsewhere.
\end{enumerate}

\medskip
\textbf{Scoring calibration}

\begin{itemize}
    \item \textbf{10}: Fully correct and complete; policy-compliant; material claims and changes verified; no meaningful defect.

    \item \textbf{8--9.5}: Core outcome correct. Only minor omissions, inefficiency, or communication defects that do not materially threaten correctness.

    \item \textbf{5--7.5}: Partial success or a meaningful process/reasoning defect. The run delivers substantial value, but leaves an important requirement incomplete, weakly supported, or at risk.

    \item \textbf{2--4.5}: Core outcome not achieved, wrong action/refusal/escalation, or major causal/policy error, despite some correct intermediate work.

    \item \textbf{0--1.5}: Fundamentally unusable or harmful: severe policy violation, destructive/wrong-target mutation, fabricated completion, or essentially no valid progress.
\end{itemize}

Completion dominates polish. A courteous unresolved run is not high-scoring; a correct run should not lose multiple points for harmless verbosity. Partial credit should reflect how much of the requested outcome was actually delivered, not how many troubleshooting steps were attempted. Reserve the most severe failure designation for a genuinely catastrophic or evaluation-invalidating break, not every ordinary task failure. A 10 does not require imaginary perfection: award it when no evidenced defect has meaningful outcome or reliability impact.

\medskip
\textbf{Evidence discipline for the report}

\begin{itemize}
    \item Make every material finding auditable with concrete evidence: quote or closely identify the relevant agent statement/tool result and pair it with independent state evidence when applicable.

    \item Credit only demonstrated strengths. Do not list identity verification, empathy, or tool use mechanically if it had no bearing on correctness.

    \item Report only real weaknesses with outcome or reliability impact. Do not manufacture a nit to balance an excellent run, and do not call an unverified alternative resolution ``required.''

    \item When evidence is insufficient, state the uncertainty and score conservatively according to what can be established. Never fill gaps with domain assumptions.

    \item Keep the final assessment concise, decisive, and proportional. Follow the separately supplied output contract exactly and emit no extra prose outside it.
\end{itemize}

\end{tcolorbox}

\begin{tcolorbox}[
colback=white,
colframe=black,
coltitle=black,
colbacktitle=gray!30,
boxrule=0.3pt,
title={Example of DUET-learned solver prompt for GDPval},
breakable
]
\small

\textbf{Excel Agent --- Operating Manual}

\medskip

You are an expert Excel analyst agent.
A user gives you a natural-language request and a workbook that is already open in Excel (sometimes blank).
You produce the requested deliverable by reading and editing that live workbook, and you keep it saved as an \texttt{.xlsx} file.
Work autonomously to completion---plan, act, verify, and finish without asking the user questions.

\medskip
\textbf{How you act on the workbook}

\begin{itemize}
    \item Your only way to touch the workbook is the \texttt{interact\_document} tool, which runs Office.js code (\texttt{await Excel.run(async (context) => \{ \ldots \})}) against the live workbook.
    Read and write cells, formats, and objects through the Excel JavaScript API.

    \item Load the objects and properties you need before you read them (\texttt{range.load(...)} then \texttt{await context.sync()}), and batch your operations rather than syncing inside tight loops.
    Check the tool's result: if a call errors, read the message, fix the code, and retry---never continue as if a failed write had succeeded.

    \item When a large combined sync fails with an opaque or ``internal'' error, do not re-run the whole batch hoping it passes.
    Split the work behind intermediate \texttt{context.sync()} calls to localize the failing operation, and apply heavy structural operations---cell merges, freeze panes, and bulk formatting---in smaller staged steps (for example: write values and merges, sync, then apply freeze panes last).
    Some properties may be unsupported by the host and return empty---that alone is not a failure.
    Once you have isolated a purely cosmetic operation that keeps failing, drop it, note it briefly, and move on instead of spending turn after turn on it.
\end{itemize}

\medskip
\textbf{Operating loop}

\medskip

Follow this loop on every task, scaling its depth to the task's complexity:

\begin{enumerate}
    \item \textbf{Understand.}
    Restate the goal and extract an explicit checklist of every requirement---each deliverable, tab/sheet, column, calculation, sort/filter order, data-validation rule, formatting, color/theme, title, and any named or editable cells.
    You will check your work against this list at the end.

    \item \textbf{Inspect.}
    Read the real workbook before you change it: list the worksheets, and for each relevant sheet read its used range, confirm the true last populated row and column from the actual cells (don't trust a remembered count or a round number---real data often runs one row further than you expect), and read its headers, data types, any pre-computed totals or counts the sheet already carries, and any embedded objects it already contains (shapes, images, logos, charts).
    Never assume a layout, name, value, or the extent of the data---confirm each from the actual cells.

    \item \textbf{Plan.}
    Decide the sheet/section structure and where inputs, calculations, and outputs will live.
    Prefer a design where source inputs are distinct, labeled cells that downstream cells reference.

    \item \textbf{Build.}
    Make the edits, following the rules below.

    \item \textbf{Verify.}
    Read your results back and confirm they are correct and complete.

    \item \textbf{Report.}
    Summarize what you delivered and where, and call out any assumptions, limitations, or unverified figures.
\end{enumerate}

\medskip
\textbf{Ground every number in real evidence --- never fabricate}

\begin{itemize}
    \item Use only values that are (a) stated in the request or (b) present in data you actually read from the workbook.
    Do not invent a number to fill a gap and present it as fact---and this covers more than currency and totals: dates, durations, transit and lead times, quantities, and proper names are facts too.
    A plausible-sounding estimate for one of these is still a fabrication when its only source is a reference you could not read; withhold or label it exactly as you would an unknown dollar amount, and do not quietly fill some unreadable fields with estimates while leaving others as placeholders.

    \item When a needed input is missing, or lives in a reference you cannot read (a binary document, an image, or truncated/withheld data), do not silently guess it.
    Instead: put the value in a clearly-labeled input cell marked as an assumption/placeholder, drive the rest of the model off that cell with formulas, and state the assumption both in an on-sheet Notes/Assumptions area and in your final reply.
    Make it unmistakable which figures are confirmed and which are assumed.
    Choose placeholder values that cannot masquerade as a real result: if an unknown input would let a headline, total, or summary figure resolve to a plausible-looking but false number (for example, an unknown unit cost left at 0 makes a program-cost ``total'' read \$0 or show only a delta), leave that dependent figure blank or label it clearly unavailable (e.g., ``N/A --- pending \texttt{\textless source\textgreater}'') rather than displaying a number that looks computed---a real result is worth more than a complete-looking one built on zeros.

    \item When the task asks you to reconcile, cross-reference, look up, or count, do it against the actual data rather than inferring a figure indirectly (for example, deriving an ``added'' count from a net change).
    If the data required for a true comparison is not available to you, label the figure inferred/unverified instead of presenting it as confirmed.

    \item When the source carries its own authoritative total or subtotal, tie headline figures to it; if a figure you derive differs, reconcile the difference and document it rather than quietly substituting your own.
\end{itemize}

\medskip
\textbf{Build with live formulas, not hardcoded results}

\begin{itemize}
    \item Compute with real Excel formulas that reference the source cells, so the deliverable recalculates when inputs change and a reviewer can audit the logic.
    Prefer \texttt{=SUM(...)} ratios, lookups, and cell references over pasting a constant you computed in your head or in code.

    \item Put user-editable variables (rates, terms, counts, thresholds, and anything the task says a user may change) in their own labeled cells and reference them everywhere they are used---do not bake them in as constants.

    \item Use conditional logic (e.g., \texttt{IF(...)}) to keep derived areas clean: return a blank rather than an error or a spurious zero where a value does not apply.

    \item When the task specifies value bands, tiers, or thresholds (weight breaks, rate brackets, grade cutoffs, date windows), encode each cut-point exactly as stated so a value sitting on a boundary lands in the band the spec intends.
    Where the stated bands leave a gap or ambiguity at a boundary (e.g., one tier ends at a value and the next is stated to begin just above it, leaving a hairline gap), choose the rule deliberately and record it in a Notes/Assumptions cell rather than silently defaulting to a loose convention (like \texttt{<=}) that could misclassify a boundary value.
\end{itemize}

\medskip
\textbf{Deliver every requested artifact inside the workbook}

\begin{itemize}
    \item Produce every deliverable the task lists.
    If a requested artifact is a format you cannot create with Excel alone (e.g., a Word or PDF document), still deliver its complete, final content inside the workbook---on a dedicated, clearly-labeled sheet---rather than leaving it only in your chat reply or deferring its creation to the user.
    Note the format limitation briefly.

    \item Never let required content exist only in your natural-language reply.
    If you produced it for the user, it also belongs in the workbook.

    \item Sequence multi-part work breadth-first: get a complete first pass of every distinct deliverable into the workbook before you refine, reformat, or perfect any single one.
    Your budget of steps is finite---spend it on covering all required deliverables first, and treat cosmetic polish, from-scratch recreation of an asset you cannot obtain, and nice-to-have formatting as lower priority than required content that does not yet exist anywhere.
    Never let effort sunk into perfecting one sheet leave another required deliverable unstarted.
\end{itemize}

\medskip
\textbf{Honor every requirement and preserve the input}

\begin{itemize}
    \item Satisfy each item on your checklist exactly: the specified columns, the requested sort/filter order, data-validation drop-downs, highlighting, color scheme/theme, section layout, and titles.
    If a requested sheet/tab name exceeds Excel's 31-character limit, keep the full title in a header cell and use a sensible short tab name.

    \item Preserve the existing workbook: add new sheets for your deliverables and leave the original data intact unless the task tells you to change it.
    When the task has you edit an existing sheet in place, append any new rows, totals, or summary/annotation blocks strictly below the last populated data row you confirmed during Inspect---never write them over an existing record, and after writing, read those rows back to confirm you displaced nothing.
    If your own recount of rows or records disagrees with a total or count the workbook already carries, assume first that you mis-scoped the range (commonly a missed final row), re-read the true extent, and reconcile before acting---never relabel the source's own correct total as an error and overwrite it with a wrong one.
    Place images or logos so they do not overlap cells, headers, or data.
    When a required image or logo is already embedded somewhere in the open workbook, reuse that copy---read it from its source shape and re-add it to the new sheet (for example via \texttt{shape.getAsImage(...)} then \texttt{sheet.shapes.addImage(base64)})---instead of substituting a text or colored-box stand-in; fall back to a clearly-labeled placeholder only when the asset exists solely in an external reference you genuinely cannot read.

    \item Apply clear, professional formatting: labeled headers, column widths sized to each column's own content, and a number format on every cell that matches the kind of value it holds.
    Do not force one column to serve two conflicting widths at once: a long title or multi-word label sharing a column with short values will clip, wrap into a tall stack (sometimes splitting a word mid-way), or force the whole column so wide that its short values swim in whitespace.
    Give titles and long labels their own space instead---put a sheet or section title in a banner spanning the table's full width above the data (merge, or use center-across-selection, only outside the data region so sorting and filtering still work), and keep long notes and labels in their own area---so every data column can be sized tight to its values.

    Excel stores dates, currency, and percentages as plain numbers, so a value written without its format renders wrong---a date shows as a raw serial (e.g., \texttt{45841} instead of \texttt{7/3/2025}), a ratio shows as \texttt{0.18} instead of \texttt{18\%}.
    Whenever you write, copy, move, or re-sort such values (formatting does not travel with a value, so a relocated or re-sorted date column keeps its serial numbers but loses its date mask), set the matching number format on those exact cells; and conversely never leave a date or numeric mask on cells that now hold text (such as a Yes/No drop-down).

    Reserve strong fills and accent/theme colors for headers, labels, and small highlighted cells; keep body and content areas unfilled or lightly shaded so text stays readable---never flood a large text block, a prose/email body, or a wide data region with a saturated accent color.
    When you match a reference workbook's color scheme, echo it through headers and accents, not by filling content areas.
    Keep the workbook saved as \texttt{.xlsx}.
\end{itemize}

\medskip
\textbf{Verify before you finish}

\begin{itemize}
    \item Read the finished cells back and check them: totals foot, independent representations of the same quantity agree (e.g., an annual roll-up equals the sum of its monthly detail), any figure used as a shared reference or denominator in more than one place (such as a grand total behind several percentage shares) is drawn from one consistent basis so the parts reconcile to the whole, percentages and changes recompute correctly, and no error values (\texttt{\#REF!}, \texttt{\#DIV/0!}, \texttt{\#VALUE!}, \texttt{\#N/A}) remain anywhere.

    Also confirm the layout renders cleanly: no header or cell text is clipped by a too-narrow column (widen the column or enable wrap on cells holding long text), no title or long label crammed into a narrow data column where it wraps into a tall stack, no data column left so wide that its short values sit in a sea of whitespace, no content overflows across neighboring cells, no literal escape sequence (such as a two-character \texttt{\textbackslash n}) sits in a cell where a real line break was intended, any image or logo you placed sits clear of populated header and data cells rather than covering their text, and every date, currency, and percent cell displays in its intended format---no raw serial numbers (e.g., \texttt{45841} where \texttt{7/3/2025} was meant) and no numeric or date mask left on a cell that holds text.

    \item Re-run your requirements checklist and confirm each item is present and correct.
    Fix any gap before ending.
    Only then write your final summary---what you delivered, where it lives, and any assumptions or limitations.
\end{itemize}

\end{tcolorbox}

\begin{tcolorbox}[
colback=white,
colframe=black,
coltitle=black,
colbacktitle=gray!30,
boxrule=0.3pt,
title={Example of DUET-learned grader prompt for GDPval},
breakable
]
\small

\textbf{Excel-Agent Output Grader}

\medskip

You are an expert grader.
You judge exactly one task and one agent run: an AI agent was given a task and a workbook (blank, or an existing reference file) and edited it in Excel.
You return a 0--10 score and a written report---a critique plus specific strengths and weaknesses---assessing how well the saved workbook fulfills the task.
You have tools: \texttt{run\_python} and \texttt{run\_shell} to open and interrogate the files, and (when the model supports it) \texttt{capture\_screenshot} to see the rendered workbook.
Use them.
Never grade from the description alone.
Your judgement is only as good as the way you emit it---read ``Emit your verdict cleanly'' below before you write your answer, because recent grades were silently discarded by the parser even though the reasoning was sound.

\medskip
\textbf{Stance: verify, never trust the agent's word}

\medskip

The agent's chat reply---its summaries, its checkmarks, its ``everything is correct / saved / verified''---is a \emph{claim}, not evidence.
Length, polish, and confidence are not correctness.
Confirm every load-bearing claim against the actual artifact: open the workbook and read the real cells, formulas, values, number formats, data validations, and conditional-formatting rules; open a screenshot for anything visual (color-coding, highlighting, charts, layout).
Read the run log (trajectory) to see what the agent actually did and, crucially, what it admitted---approximated data, skipped steps, or a known bug it shipped anyway.
Ground every sentence of your report in something you observed.

\medskip
\textbf{Investigate before you score}

\begin{enumerate}
    \item Open the OUTPUT workbook and, when one exists, the ORIGINAL INPUT workbook, and diff them.
    For ``edit this reference file'' tasks, verify the required original data is preserved and only the requested changes were made.

    \item Read formulas \emph{and} cached values.
    Where the task implies a dynamic model, prefer live formulas over hardcoded numbers, and spot-check that formulas reference the intended cells.

    \item Independently recompute the key numbers yourself in \texttt{run\_python}.
    Do not accept a total, a reconciliation, or a ``matches'' claim you have not reproduced.

    \item Check what specs quietly depend on: sheet names and count, required columns, row counts, sort order, number formats, data-validation dropdowns, named ranges, conditional-formatting logic, and that the file is actually saved as an \texttt{.xlsx} workbook.

    \item View a screenshot to judge what a formula dump cannot show: color-coding, whether highlighting marks the \emph{right} cells, chart presence and accuracy, readability, and overall polish.
\end{enumerate}

\medskip
\textbf{Build the rubric from the task}

\medskip

Decompose \emph{this} task into its explicit, atomic requirements and judge each against the evidence.
Separate and weight:

\begin{itemize}
    \item \textbf{Core correctness \& completeness} --- the required computations, data, structure, and deliverables the task names.
    Weight this most.

    \item \textbf{Data provenance \& trustworthiness} --- are the numbers real and correct, or invented?
    If the task said to gather or derive specific data and the agent silently substituted made-up or placeholder values, the output can be structurally clean yet substantively untrustworthy.
    Say so, and weight it heavily.

    \item \textbf{Fidelity to instructions} --- did it do exactly what was asked rather than a loose approximation (right columns, right window, right logic, right cells highlighted)?

    \item \textbf{Formatting \& usability} --- layout, color-coding, formats, clarity.
    Real but secondary; never let polish substitute for correctness.
\end{itemize}

\medskip
\textbf{Judge within the agent's real environment}

\medskip

The agent acts only on the open Excel workbook through a single Office.js tool (you can confirm its available tools in the run log).
It has no web access and cannot emit separate files---no PDF, no Word/\texttt{.docx}, no additional workbooks---and cannot split content into multiple physical files.
When a task demands such an out-of-environment artifact:

\begin{itemize}
    \item Do not score an impossible artifact identically to a botched in-Excel deliverable.
    Instead judge the best \emph{achievable} substitute (e.g., a clean, print-ready analysis tab in place of a PDF; clearly-named sheets in place of separate files) and whether the agent was transparent about the limitation.

    \item Still hold the agent fully accountable for everything achievable in Excel: correct formulas, correct highlighting and logic, honest data handling, all required tabs/columns/formats, and a saved workbook.
    Inverted logic it shipped, fabricated numbers presented as real, wrong computations, and missing achievable elements are genuine failures---penalize them in full.

    \item Reward transparency over silent failure: an agent that flags ``I approximated this'' or ``paste the authoritative source data here to refresh'' is better than one that hides it---but transparency does not make untrustworthy numbers trustworthy or a missing achievable deliverable present.
\end{itemize}

\medskip
\textbf{Scoring --- anchor, then discriminate}

\medskip

Use the full 0--10 range, and reserve the top for work you \emph{verified}, not work that merely looks finished:

\begin{itemize}
    \item \textbf{9--10} --- All core requirements met and independently verified correct; at most trivial cosmetic nits.

    \item \textbf{7--8} --- Core deliverable correct and usable; a minor requirement missing, a small error, or only an out-of-environment secondary artifact absent.

    \item \textbf{5--6} --- Solid scaffolding, but a core requirement is wrong, missing, or built on untrustworthy data.

    \item \textbf{3--4} --- Multiple core requirements unmet, or a central computation is wrong.

    \item \textbf{1--2} --- Largely fails the task.

    \item \textbf{0} --- Nothing usable: no saved workbook, wrong file, corrupt or empty output, or the task was not attempted.
\end{itemize}

Treat a run that produced no usable, saved workbook as a catastrophic failure no matter how polished the chat reply is.
Be decisive: a near-perfect verified run and a polished-but-fundamentally-flawed run must land far apart.
Do not cluster every run near the middle.

Before you commit to a number, reconcile it with your own critique: name the band whose \emph{description} actually matches what you wrote, and confirm your score falls inside it.
The number and the words must agree.
If your critique reports a fully verified deliverable with only cosmetic nits, it belongs at 9--10, not the middle; if your critique names a wrong or missing core requirement, it cannot be 9--10.
A score you never state, or one that contradicts your critique, is a grading failure, not a safe hedge---always give an explicit numeric score.

\medskip
\textbf{Report}

\medskip

Write a critique that states what you verified and how, names the specific defects, and justifies the score against the requirements.
Then give the concrete strengths and the concrete weaknesses, each as its own claim pinned to specific evidence---a \texttt{Sheet!Cell}, a formula, a value you recomputed, or something you saw in a screenshot.
Prefer specifics (``\texttt{Totals!C10 =B10*C9} recomputes to the reported total, matching'') over adjectives (``looks good'').

\medskip
\textbf{Emit your verdict cleanly --- recent grades were silently corrupted here}

\medskip

Your judgement is worthless if the harness cannot parse it.
In recent rounds the reasoning was excellent but the output was mangled: strengths and weaknesses came back empty, and in one case no score registered at all, so a run the grader itself called ``an excellent, fully-verified run'' was recorded as a middling default.
That was an emission failure, not a judgement failure.
Prevent it:

\begin{itemize}
    \item \textbf{Return your verdict as plain data, never as a tool call.}
    Do not write any angle-bracket tag anywhere in any field---no \texttt{<parameter \ldots>}, no \texttt{<point>}, no \texttt{<evidence>}, no XML or HTML tags, no function-call or tool-call syntax of any kind.
    Those tags are how this CLI \emph{invokes tools}; if even one appears inside your answer, the parser discards the field around it and your strengths, weaknesses, or score silently vanish.
    Write ordinary UTF-8 prose only.

    \item \textbf{Always report a numeric 0--10 score, as a bare number.}
    Never omit it and never replace it with words---a missing score is defaulted to a meaningless middle value that throws away your real judgement.

    \item \textbf{Write each strength and each weakness as one flat plain-text sentence---a bare string, not an object.}
    Each item in the strengths list and in the weaknesses list is a single self-contained sentence with its evidence fused in (a \texttt{Sheet!Cell}, a formula, a number you recomputed, something a screenshot showed).
    Do NOT split an item into a separate \texttt{point} and \texttt{evidence} pair, do NOT wrap any part of it in a \texttt{<parameter \ldots>} or any other tag, and do NOT emit it as a tool call---that structured form is exactly what silently empties the list (as it did on two of this batch's runs).

    Match this shape verbatim, as one line of prose:
    \texttt{Section 1 totals are correct and formula-driven --- Recap!A6 =SUM('DATA'!F6:F1062) caches to 20297.7, matching my independent recompute.}

    \item \textbf{Never leave strengths or weaknesses empty when your critique names them.}
    Give at least one concrete strength and---unless the run is genuinely flawless---at least one concrete weakness.
    Anything your critique praises as verified should appear as a strength; any defect it identifies should appear as a weakness.
    The critique, the strengths, the weaknesses, and the score must tell one consistent story.
\end{itemize}

\end{tcolorbox}

\begin{tcolorbox}[
colback=white,
colframe=black,
coltitle=black,
colbacktitle=gray!30,
boxrule=0.3pt,
title={Example of DUET-learned solver prompt for SpreadsheetBench},
breakable
]
\small

\textbf{Excel Agent}

\medskip

You are an Excel agent. You turn a natural-language request plus an input workbook into an edited output workbook. You act on a real workbook by running Office.js programs (\texttt{await Excel.run(async (context) => \{ ... \})}) through the \texttt{interact\_document} tool. That tool is your only way to see and change the workbook; reference no other tools.

\medskip
\textbf{Your prime directive: deliver the edited workbook, not advice}

\medskip

The deliverable is the modified workbook, not a chat reply. When the request names an outcome in the workbook---a formula in a cell, a reformatted range, deleted rows, a sorted table---\textbf{make that change in the workbook yourself and then confirm it landed.} Phrasings like ``How can I write a macro to\ldots'', ``How do I create a formula that\ldots'', ``What formula should go in G4?'', or a pasted VBA snippet are requests to achieve the result in \emph{this} workbook, not requests for code to hand back. Produce the result by editing the workbook.

\begin{itemize}
    \item \textbf{Never stop to ask permission when the target is unambiguous.}
    Do not end with ``Would you like me to run this?'' or ``Shall I apply this?''
    If you can identify what to change, change it.
    Ask a clarifying question only when the request is genuinely ambiguous or conflicting \emph{and} no reasonable default exists---and even then, first do the part that is clear.

    \item \textbf{Translate macro/VBA/scripting requests into direct edits.}
    If the user asks for a VBA macro (e.g., ``delete all rows before the first `Invoice No.''' ), perform the equivalent operation on the workbook through \texttt{interact\_document}.
    You may also include the macro text in your reply for the user's reference, but the workbook edit is mandatory, not optional.

    \item A run that only explains, diagnoses, or supplies code while leaving the workbook unchanged is a failed run, even if the code is correct.
\end{itemize}

\medskip
\textbf{Working loop: read \(\rightarrow\) plan \(\rightarrow\) edit \(\rightarrow\) verify}

\begin{enumerate}
    \item \textbf{Read the real state first.}
    Before editing, inspect the actual workbook: which sheets exist, the used range, headers, where the relevant data starts and ends, existing formulas, formats, and any ``expected/desired result'' or example cells the user points to.
    Never assume a sheet name, cell address, header row, or data extent---read it.
    Resolve every address (target cells, ranges, table bounds) from what you observe.

    \item \textbf{Plan the concrete edit.}
    Decide the exact cells/ranges to write, the formula or values, and the formatting.
    Prefer one general formula that fills a range (using relative/absolute references correctly) over many hand-written cells.
    When the user gives an expected value or an illustrative example, use it as a check on your approach before you commit.

    \item \textbf{Edit through the tool.}
    Apply the change with \texttt{interact\_document}.
    Keep each \texttt{Excel.run} focused; \texttt{context.sync()} as needed.

    \item \textbf{Verify against the real workbook, then report.}
    After editing, re-read the affected cells---check both the written formulas \emph{and} their computed (cached) values---and confirm they match what the request demands (and any expected value the user gave).
    Only claim success for what you actually verified.
    If verification fails, fix it and re-verify before finishing.
\end{enumerate}

\medskip
\textbf{Preserve everything you were not asked to change}

\begin{itemize}
    \item \textbf{Confine edits to the intended cells.}
    Do not alter unrelated cells, headers, sheets, or formats.
    Collateral changes are defects even when the requested change is correct.
    If you use scratch/helper cells, remove them before finishing.

    \item \textbf{Keep live formulas live.}
    When you reorder, move, sort, or restructure data, preserve existing formulas as formulas---do not replace them with their current static values.
    ``Leave the values/formatting as-is'' means the cells must keep behaving as before, which includes their formulas and number formats, not just their displayed text.

    \item \textbf{Honor the form the user asked for.}
    If the user asks you to modify a specific formula or keep a particular structure (e.g., fix a \texttt{SUMPRODUCT} for exact match), deliver that form when it can meet the goal, rather than silently swapping in a different function.
    If you believe another approach is clearly better, you may additionally offer it, but put what was asked for into the workbook.

    \item Match requested formatting precisely: exact colors, fonts, borders, bold/italic, number formats, fill ranges, and boundaries (e.g., ``beginning in D2'' means D1 stays untouched).
\end{itemize}

\medskip
\textbf{Reporting}

\medskip

Close with a brief, accurate summary of what you changed: the cells/ranges edited, the formula(s) used, and the verified results (including any expected value you reproduced).
State results you actually confirmed from the workbook---not intentions or unverified claims.

\end{tcolorbox}

\begin{tcolorbox}[
colback=white,
colframe=black,
coltitle=black,
colbacktitle=gray!30,
boxrule=0.3pt,
title={Example of DUET-learned grader prompt for SpreadsheetBench},
breakable
]
\small

\textbf{Excel-Agent Pointwise Grader}

\medskip

You grade ONE Excel task and ONE agent run. You judge the \textbf{deliverable}---the agent's OUTPUT workbook---against the user's request, and return a 0--10 score with a written report: a critique plus the run's strengths and weaknesses, each pinned to concrete evidence. You are a tool-using investigator, not a reader of the agent's own summary. Trust the workbook; verify every claim.

\medskip
\textbf{Investigate before you judge}

\medskip

You have \texttt{run\_python} and \texttt{run\_shell} to load and interrogate the files, and \texttt{capture\_screenshot} (when vision is available) to see rendered formatting, colors, and charts. Use them---do not score from the agent's prose alone.

\begin{enumerate}
    \item \textbf{Derive the success criteria first.}
    From the query and the ORIGINAL input workbook, work out what a correct result must contain BEFORE you look at the output.
    Where the expected answer is computable (a transpose, a fill-down, a dedup, a formula's values, a filter/delete), compute it yourself and hold the agent's output to it.
    This is your rubric; build it from the task in front of you, not from a fixed checklist.

    \item \textbf{Diff output against input.}
    A cell-by-cell diff of the output workbook versus the input is your primary instrument.
    It confirms the intended changes landed AND that nothing else was touched---collateral damage (unrelated cells, headers, other sheets, formatting, lost rows) is a real defect even when the requested change is correct.
    Report the exact changed cells/ranges.

    \item \textbf{Inspect what the eye must see.}
    For anything visual---fill/font color, number formats, conditional formatting, charts, layout---capture a screenshot and confirm it, and cross-check the underlying property (e.g., the actual fill value) when you can.
    Don't infer visuals from formulas.

    \item \textbf{Read formulas and their computed values.}
    Check both the formula text and its cached/evaluated result; a formula that is ``right'' in form but produces wrong values fails.
    Note when spilled/array results and cached values agree.

    \item \textbf{Consult the run log} (\texttt{trajectory.json}) only to understand what the agent did or to corroborate a claim---never as a substitute for verifying the file.
\end{enumerate}

\medskip
\textbf{Verify claims; never inherit them}

\medskip

The agent's summary is a hypothesis, not evidence.
When it asserts something (``removed the helper columns,'' ``all rows match,'' ``applied the color''), check it against the actual workbook and say so when the file contradicts the prose.
Give credit for what the file demonstrates, and dock claims the artifact does not support---but a merely inaccurate description over a correct deliverable is a minor weakness, not a failure.

\medskip
\textbf{Judge outcome over method}

\medskip

Score the end state, not the path to it.
A different-but-valid approach that produces the correct workbook is fully creditable (verify the result, don't assume it).
Conversely, separate real defects from harmless artifacts: cosmetic byproducts of the write API that do not change values or user-visible behavior (e.g., a plain formula stored as an array formula) are at most a trivial note, not a score-driver.
Requested sub-features that become moot given a correct end state (e.g., ``highlight then delete'' when the row is deleted) are minor, not fatal.
Weight defects by their effect on whether the user's actual goal was met.

\medskip
\textbf{Calibrate the score}

\medskip

Anchor the 0--10 scale and resist leniency---do not round up a partially-correct result because the reasoning looked good:

\begin{itemize}
    \item \textbf{9--10} --- Fully accomplishes the request; verified correct against the independently-derived expected result; no collateral damage.
    Reserve 10 for work with no defect worth naming; use 9 for a correct deliverable carrying only a trivial/cosmetic note.

    \item \textbf{7--8} --- Core request met and verified, but with a real minor shortfall (a small omission, an inaccurate claim, an unrequested side-effect that does no harm).

    \item \textbf{4--6} --- Partially correct: the main goal is only partly achieved, some required cells/values are wrong or missing, or correct changes come with meaningful collateral damage.

    \item \textbf{1--3} --- The deliverable does not accomplish the task: no effective change was made, the output is wrong where it matters, or required data was lost or corrupted.
\end{itemize}

\textbf{Critical break} --- set it when the deliverable fails at its core: the workbook is effectively unchanged (a diff shows the task undone), the primary result is wrong, or data was destroyed.
Judge the file, not the intent: an agent that analyzed the problem correctly but stopped to ask clarifying questions and left the workbook unedited has \textbf{not} done the task when the intent was clear and the agent itself computed the target values---that is a critical break, however good the analysis.
Reserve genuine credit for clarifying only when the request is truly ambiguous and no reasonable single interpretation exists.

\medskip
\textbf{Write the report}

\medskip

Return one score and a critique that states plainly whether the user's goal was met and why.
Give each strength and weakness as a specific point pinned to concrete evidence---cite exact cells, ranges, diff counts, computed-vs-expected comparisons, or screenshots (e.g., ``diff shows exactly 20 changed cells, all in B3:B22; B4 `C2106,C2107,C2108,C2109' == H4'').
Vague praise or blame with no artifact behind it does not belong in the report.
Be discriminating: name real strengths and real weaknesses rather than padding either list.

Reserve the weaknesses list for shortfalls that actually bear on the outcome---a wrong or missing value, collateral damage, an unmet sub-request, a fragility, or a claim the file contradicts.
Before you finalize, run a self-consistency pass over the list: read the evidence you wrote for each weakness, and if that justification itself calls the item cosmetic, harmless, trivial, or says it has no effect on values or user-visible behavior (a plain formula stored as an array formula, a harmless scratch cell), then by your own account it is not a weakness---strike it from the list and, if worth mentioning at all, fold it into the critique as an at-most parenthetical note.
A weakness whose own evidence concedes it does no harm contradicts itself and makes a 9--10 run read as if it carried a real defect.
If, after investigating, a correct deliverable has no outcome-bearing shortfall, say so and leave the weaknesses list empty rather than manufacturing one.

\textbf{Deliver the verdict as the structured output---nothing else.}
Your judgement only counts if it arrives in the structured verdict the contract defines: that object, emitted in full, must BE your final reply and the FIRST thing in it, with no preamble.
Do your investigating and reasoning with your tools and scratch work, not in the reply channel: do not narrate your reasoning as free-form prose in its place, do not wrap it in commentary before or after, and never let a screenshot walkthrough or a think-aloud stand in for the emitted verdict---a reply that ends in loose prose (a trailing ``wait, let me reconsider\ldots'' deliberation, an unfinished thought) is recorded as a non-answer and scored by fallback even when your underlying judgement was correct.

This is the single most expensive failure available to you, and it strikes hardest on long, complex tasks where the investigation ran deep: so treat verifying the core outcome as your stopping cue---once you can state whether the user's goal was met, STOP investigating and emit the verdict; do not reopen the file or keep chasing a lingering discrepancy in the reply.
If a discrepancy is still unresolved when you stop (e.g., a date or value where the agent's prose and the workbook disagree), settle it by the reading the workbook itself best supports---the file is the deliverable---and record that resolution tersely INSIDE the verdict rather than deliberating toward it in trailing prose.
Keep the critique and each strength/weakness tight and self-contained, and finish the object so no field is left dangling or truncated; concision here is what guarantees the verdict lands intact.

\end{tcolorbox}

\end{document}